\RequirePackage{float}
\documentclass[a4paper,fleqn]{cas-sc}

\usepackage[T1]{fontenc}
\usepackage[utf8]{inputenc}
\usepackage{newunicodechar}
\newunicodechar{–}{\textendash}
\newunicodechar{—}{\textemdash}
\usepackage{amsmath,amssymb}
\usepackage{graphicx}
\usepackage{booktabs,multirow,arydshln}
\usepackage{pifont}
\usepackage{url}
\usepackage{lineno}
\usepackage{etoolbox}
\AtBeginEnvironment{abstract}{\selectfont}
\AtBeginEnvironment{table}{\small}
\AtBeginEnvironment{figure}{\selectfont}
\newsavebox{\csftablebox}
\newcommand{\csffittable}[1]{%
  \sbox{\csftablebox}{\fontsize{9}{11}\selectfont#1}%
  \ifdim\wd\csftablebox>\linewidth
    \resizebox{\linewidth}{!}{\usebox{\csftablebox}}%
  \else
    \usebox{\csftablebox}%
  \fi}

\hypersetup{hidelinks}
\makeatletter
\newcommand{\csfappendixseccntformat}[1]{%
  \ifstrequal{#1}{section}{Appendix~\thesection.\quad}{\csname the#1\endcsname\quad}}
\apptocmd{\appendix}{%
  \renewcommand{\thesection}{\Alph{section}}%
  \let\@seccntformat\csfappendixseccntformat
}{}{}
\makeatother
\newcommand{\cmark}{\ding{51}}
\newcommand{\xmark}{\ding{55}}
\usepackage[numbers,compress]{natbib}
\shorttitle{Latent Distribution Modeling for Source Localization}
\shortauthors{D. Hou et al.}

\renewcommand{\ttdefault}{cmtt}

\begin{document}

\title[mode=title]{Latent Distribution Modeling for Source Localization under Training-Inference Observation Mismatch}

\author[aff2,aff1]{Dongpeng Hou}
\ead{hdp0918@mail.nwpu.edu.cn}
\author[aff2]{Junqing Zhao}
\ead{zhaojunqing@mail.nwpu.edu.cn}

\author[aff2]{Chao Gao}
\ead{cgao@nwpu.edu.cn}

\author[aff2]{Xianghua Li}
\ead{li_xianghua@nwpu.edu.cn}

\author[aff4]{Zhen Han}
\ead{hanzhen@nwpu.edu.cn}

\author[aff5,aff3]{Andrey Leonidov}
\ead{leonidovav@lebedev.ru}

\author[aff3]{A. M. Raigorodskii}
\ead{MRaigor@yandex.ru}

\affiliation[aff2]{organization={School of Artificial Intelligence, Optics and Electronics, Northwestern Polytechnical University},
  city={Xi'an}, postcode={710072}, country={China}}
\affiliation[aff1]{organization={School of Mechanical Engineering, Northwestern Polytechnical University},
  city={Xi'an}, postcode={710072}, country={China}}
\affiliation[aff4]{organization={Hangzhou International Innovation Institute, Beihang University},
  city={Hangzhou}, postcode={311115}, country={China}}
\affiliation[aff5]{organization={Lebedev Physical Institute of the Russian Academy of Sciences},
  city={Moscow}, postcode={119991}, country={Russian Federation}}
\affiliation[aff3]{organization={Moscow Institute of Physics and Technology},
  city={Dolgoprudny}, state={Moscow Region}, postcode={141701}, country={Russian Federation}}

\begin{abstract}
Source localization seeks to infer the origins of network propagation processes, with applications to tracing misinformation in social networks. Recent discriminative methods exploit temporal propagation sequences to directly predict source probabilities, but models trained on such sequential observations typically require temporal propagation sequences to remain available during inference. However, in practice, although complete sequences can be used for offline training, collection delays often restrict timely inference to the use of a single observed snapshot. The resulting observation mismatch induces a distributional shift between dynamics-enriched training representations and snapshot-based inference representations, degrading source localization performance. To address this challenge, we propose Temporally Guided Latent Refinement (TGLR). During training, node features are encoded into latent distributions, and temporal dynamics guide residual refinement of the resulting distributions. A latent diffusion model is then trained to broaden the resulting latent distribution and capture its structure. During inference, test-time adaptation uses the trained diffusion model as a distributional prior to adapt snapshot-based latent representations toward the dynamics-enriched distribution. We also construct Twitter25, a real-world dataset incorporating quoting interactions, to evaluate source localization on richer propagation structures. Extensive experiments across thirteen simulated and real-world propagation settings demonstrate the effectiveness of TGLR under the training–inference observation mismatch, with an average relative F1 improvement of \textbf{3.1}\% over baseline methods.
The code and datasets are available at \url{https://github.com/DancinPuppet/TGLR}.
\end{abstract}

\maketitle
\suppressfloats[t]
\hypersetup{pdfauthor={Dongpeng Hou, Junqing Zhao, Chao Gao, Xianghua Li, Zhen Han, Andrey Leonidov, A. M. Raigorodskii}}


\section{Introduction}
Online social networks enable information to diffuse rapidly across large populations, yet they also allow misinformation and rumors to propagate before they can be verified. False news has been shown to spread faster and reach more people than true news on social platforms~\cite{vosoughi2018}, amplifying its potential societal impact. For example, during the recovery efforts after Hurricane Helene in the United States, false claims about FEMA's disaster relief policies spread widely on social media, fostering public distrust toward relief workers, triggering violence, and delaying aid delivery~\cite{Rokhideh2025}.  In high-impact events such as public health crises, political campaigns, or international conflicts, source localization seeks to identify the nodes from which an observed propagation cascade originates and can provide evidence to support subsequent intervention and effective response.

\begin{figure}[pos=tbp]
    \centering
    \includegraphics[width=0.65\columnwidth, trim={0.5mm 0.5mm 0.5mm 0.5mm}, clip, ]{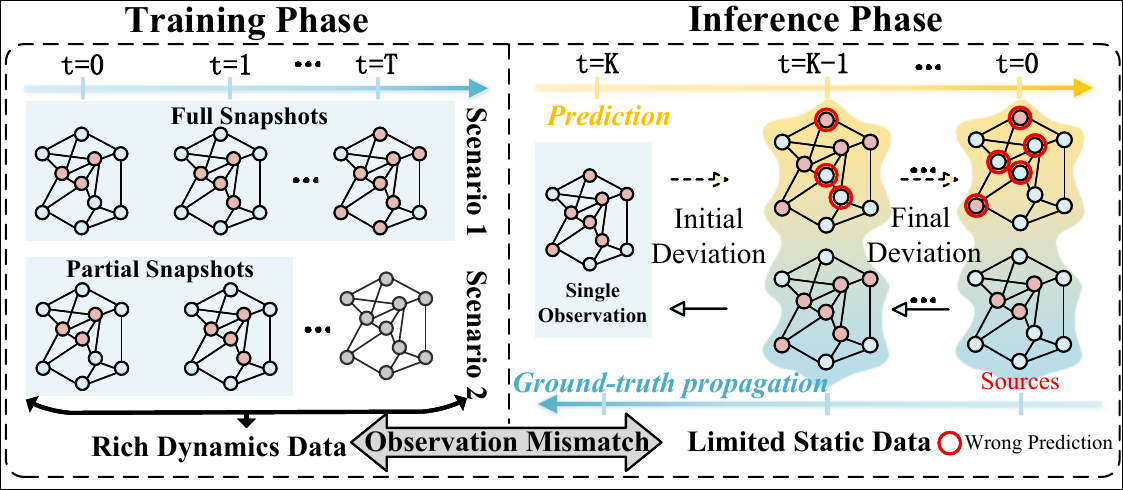}
    \caption{Illustration of the training-inference observation mismatch 
    in source localization. During training, models learn from complete 
    temporal propagation sequences with rich dynamics. During inference, 
    only a single observed snapshot is available.}
    \label{fig:motivation}
\end{figure}

Source localization is commonly studied through models of network propagation dynamics, including the Susceptible-Infected-Recovered (SIR) model and its variants~\cite{zhu2016sir, jiang2024signed, yu2024hypergraph}, the Susceptible-Infected (SI) model~\cite{kazemitabar2020source, zang2015topic}, and the Susceptible-Infected-Susceptible (SIS) model~\cite{luo2013sis}, which characterize how node states evolve over a network. Regarding the acquisition of observational data, existing approaches are generally categorized into two paradigms: those based on sensor observations and those relying on snapshot observations. Sensor-based methods utilize observer nodes to record the arrival information of the diffusion process, such as infection timestamps or propagation directions~\cite{hou2024new, ma2026rumor, cheng2022pslm, gong2024hmsl}, but their applicability can be limited by deployment and collection costs~\cite{wang2022rapid} and by the need for network topology when selecting observer locations. Snapshot-based methods instead infer sources from an observed propagation graph and its node states, avoiding dependence on predeployed sensors and their temporal records. We therefore focus on snapshot-based source localization, which matches the observation setting considered in this work.

Early snapshot-based source localization methods estimate source likelihood by iteratively propagating infection state information across the network neighborhood without requiring prior knowledge of the underlying diffusion model~\cite{wang2017LPSI}. With the development of deep learning, learning-based source localization methods have become increasingly common and can be broadly grouped into generative and discriminative paradigms following the classical modeling taxonomy~\cite{ng2001discriminative, Liu2017}.

{\bfseries Generative methods} formulate source localization as an inverse or optimization problem~\cite{IVGD22, chen22slvae}, modeling the information diffusion process to learn the underlying dynamic mechanisms~\cite{xin23DDMSL, liu2024DDSL}. These methods can provide physical interpretability~\cite{IVGD22} but some require iterative forward simulation or backward inversion through the generative process~\cite{chen22slvae, xin23DDMSL, PDSL2026, GDFSL2025}. Because the intermediate propagation states are unobserved, the inverse process is insufficiently constrained at inference time~\cite{kabanikhin2008definitions}. Small errors in the observed snapshot can therefore be amplified during source estimation, undermining the stability of localization results. These limitations motivate methods that avoid reconstructing the complete propagation process during inference.

{\bfseries Discriminative methods} typically formulate source localization as a node classification problem, learning a direct mapping from node features to source probabilities~\cite{hou2023TGASI, SIDSL2025}. Unlike generative approaches that require iterative backward inference through unobserved intermediate propagation states, discriminative methods directly predict source probabilities from observed snapshots. Early discriminative methods operate on a single infection snapshot, leveraging graph convolutional networks to aggregate structural and state features for source classification~\cite{ming19GCNSI, guo2021igcn}. To capture richer diffusion dynamics, recent methods incorporate temporal propagation sequences, designing temporal graph attention mechanisms and multi-granularity temporal signals to model the evolution of node states across successive snapshots~\cite{hou2023TGASI, lu2025GTSI}. 

However, models trained on such sequential observations typically require temporal propagation sequences to remain available during inference, implicitly assuming consistent observational conditions between training and inference. In practice, although complete temporal sequences can be used as privileged information during offline training, continuous monitoring and temporal data aggregation introduce collection delays that often restrict timely inference, so training learns dynamics-aware representations while inference is typically constrained to a single observed snapshot (Figure~\ref{fig:motivation}). This mismatch induces a distributional shift between dynamics-enriched training representations and snapshot-based inference representations, limiting their localization performance at inference time.

To address these challenges, we propose Temporally Guided Latent Refinement (TGLR). The encoder maps structural and propagation state features to a base latent distribution that can be computed from a single observed snapshot. During training, temporal propagation sequences guide residual refinement of the latent mean, yielding a dynamics-enriched distribution. Then, a latent diffusion model is trained from this refined distribution to broaden it and capture its structure, enabling the trained denoiser to serve as a distributional prior. During inference, when temporal sequences are unavailable, test-time adaptation leverages this prior to iteratively adapt snapshot-based representations toward the dynamics-enriched distribution, thereby mitigating the distributional shift. To evaluate source localization on richer propagation structures, we construct Twitter25 by incorporating quoting interactions into propagation cascades. Experiments on both real-world and simulated datasets demonstrate the effectiveness of TGLR under the training–inference observation mismatch.

The main contributions of this work are summarized as follows:
\begin{itemize}

\item We characterize the training-inference observation mismatch in discriminative source localization, where offline training exploits complete temporal propagation sequences, but timely inference is limited to a single observed snapshot. We mitigate the resulting distributional shift through latent distribution refinement, with controlled comparisons demonstrating its advantages over distribution alignment strategies.

\item We propose TGLR, a source localization method that combines temporally guided residual refinement with latent diffusion modeling. Temporally guided residual refinement incorporates propagation dynamics into latent distributions during training. Latent diffusion modeling broadens the refined latent distribution and captures its structure during training, providing a distributional prior that guides test-time adaptation of snapshot-based representations when temporal information is unavailable during inference.

\item We construct Twitter25, a real-world propagation dataset incorporating quoting interactions, to support source localization evaluation on richer propagation structures.

\item Experiments across thirteen simulated and real-world propagation settings demonstrate the effectiveness of TGLR in leveraging temporal propagation information available during training to improve source localization from a single snapshot at inference, achieving an average relative F1 improvement of \textbf{3.1}\% over baseline methods.

\end{itemize} 

\section{Related Work}

This work draws on two lines of research: source localization and
latent diffusion. The former covers methodological developments in
source localization, including how different approaches use propagation
observations for source inference. The latter learns distributional
structure in latent space through denoising, providing a methodological
basis for the distributional prior used to adapt snapshot-based
representations in our method.

\subsection{Source Localization}

The problem of source localization in network diffusion processes has been extensively studied through various computational approaches and theoretical frameworks. Early work established foundational propagation models to characterize information spreading dynamics, including epidemic models such as the SI, SIR, and SIS models. In parallel, influence propagation models, including the Independent Cascade (IC)~\cite{goldenberg2001ic} and Linear Threshold (LT)~\cite{granovetter1978lt} models, provided alternative representations for modeling the propagation of influence or information across network structures. Jiang et al.~\cite{jiang2024signed} and Yu et al.~\cite{yu2024hypergraph} develop dynamic message-passing methods that rely on predefined SIR-type propagation models for source inference in signed networks and hypergraphs, respectively.

Without requiring prior knowledge of the propagation model, Wang et al.~\cite{wang2017LPSI} propose an infection-state label propagation strategy that can operate under both epidemic and influence propagation settings. Here, infection states are iteratively propagated through the network to identify candidate sources. To leverage multi-hop neighborhood information for improved prediction, Dong et al.~\cite{ming19GCNSI} integrate these propagation-derived features into Graph Neural Networks (GNNs), enabling end-to-end learning of richer structural representations. Wang et al.~\cite{IVGD22} further formulate source localization as a graph diffusion inverse problem, enabling source inference through reverse diffusion processes.

To address the uncertainty inherent in inverse localization problems, recent work has increasingly adopted probabilistic generative models, primarily Variational Autoencoders (VAEs)~\cite{vae13} and denoising diffusion models~\cite{ddpm20}. VAE-based methods encode infection probabilities into latent distributions to quantify uncertainty. Ling et al.~\cite{chen22slvae} employ VAEs to learn latent representations for source localization, while Ling et al.~\cite{ling2024cnsl} extend this framework to multi-network scenarios where sources must be inferred across interconnected platforms. Wang et al.~\cite{PDSL2026} further develop a VAE-variant source quantification module conditioned on propagation dynamics and use Graph Neural ODEs to model continuous node-state evolution for inference refinement.

Diffusion-based methods model uncertainty through denoising or reverse diffusion processes. Liu et al.~\cite{liu2024DDSL} combine an invertible GNN with diffusion-based source generation under snapshot observations, while Yan et al.~\cite{xin23DDMSL} model information diffusion via Markov chains and use a reversible residual network for source localization and path reconstruction. Hou et al.~\cite{GDFSL2025} reformulate the forward diffusion process to encode propagation-specific noise and derive a closed-form reverse process for source inference across diverse diffusion mechanisms. However, relying on generative inversion or iterative simulation for inference can be risky in ill-posed scenarios. Without sufficient constraints, these methods are prone to error amplification, where minor perturbations in observations destabilize the localization results.

To avoid the instability introduced by generative inversion, discriminative methods directly estimate source probabilities from observed graph states. Under the single-snapshot setting, Chen et al.~\cite{SIDSL2025} construct topology-aware priors from graph label propagation and use them to guide source distribution estimation from graph topology and observed node states. To incorporate propagation dynamics beyond the structural and state information available in a single snapshot, recent discriminative approaches use temporal propagation sequences to enhance node discriminability. Hou et al.~\cite{hou2023TGASI} formulate source localization as a sequence-to-sequence problem using GNN-based encoding and temporal attention for inductive inference under time-varying propagation. Lu et al.~\cite{lu2025GTSI} introduce a Graph Transformer architecture with propagation-label-based attention bias to jointly model static topology and dynamic infection states, enabling global dependency modeling for source localization under time-varying propagation. However, these methods often presume similar observational conditions during training and inference. In practice, training can exploit richer temporal propagation information, whereas inference is typically limited to a single observed snapshot. This observation mismatch induces a distributional shift between training and inference representations, limiting source localization performance.

\subsection{Latent Diffusion}

Recent advances in generative modeling have explored both VAEs and DDPMs as core architectures. In the graph generation domain, mainstream paradigms include variational autoencoder–based methods~\cite{graphvae18}, diffusion-based methods~\cite{digress22, jo2022score}, auto-regressive models~\cite{graphrnn18}, generative adversarial networks~\cite{maziarka2020molcygan}, and normalizing flow models~\cite{moflow20, luo2021graphdf}, which aim to learn expressive graph representations capable of producing realistic and structurally coherent graphs.

Beyond single-paradigm designs, the integration of VAEs and diffusion processes in a shared latent space has given rise to a new research direction—latent diffusion. In the image domain, Rombach et al.~\cite{ldm22} introduce latent diffusion models that deliver strong performance across diverse generation tasks while being more computationally efficient than pixel-space diffusion. Inspired by these advantages, recent work has adapted the latent diffusion paradigm to graphs. Yang et al.~\cite{graphusion24} map graphs to a low-dimensional latent space via a variational graph autoencoder, apply noise injection, and perform SDE-based~\cite{song2020sde} denoising for generation, achieving smoother and more expressive results but still incurring notable runtime overhead due to stepwise noise inversion. In TGLR, latent diffusion modeling learns the structure of dynamics-enriched latent distributions and provides a distributional prior for adapting snapshot-based representations during inference.

\section{Methodology}

\subsection{Problem Formulation}

Consider an undirected graph $\mathcal{G} = (\mathcal{V}, \mathcal{E})$, where $\mathcal{V} = \{v_1, v_2, \ldots, v_n\}$ denotes a node set and $\mathcal{E}$ denotes an edge set representing propagation
relationships between nodes. The adjacency matrix $A \in \{0,1\}^{n \times n}$ is defined by
$A_{i,j} = \mathbb{I}[(v_i, v_j) \in \mathcal{E}]$, where $\mathbb{I}(\cdot)$ denotes the indicator function.
A set $\mathcal{S} \subseteq \mathcal{V}$ represents the source nodes that initiate a propagation process.
The diffusion unfolds over discrete time steps, and the observed infection status at time $t$ is represented by a binary vector
$\mathbf{Y}_t \in \{0,1\}^n$, where $Y_{t,i} = 1$ indicates that node $v_i$ is currently infected, and
$Y_{t,i} = 0$ otherwise. Under the SIR model, both susceptible and recovered nodes are encoded as 0.
At inference, the available propagation observation is a single snapshot $\mathbf{Y}_T$ at an observation time $T$. The goal of source localization is to identify the source set $\hat{\mathcal{S}}$ that maximizes the posterior probability given the graph and the observation:
\begin{equation}
\hat{\mathcal{S}} = \arg\max_{\mathcal{S} \subseteq \mathcal{V}} P(\mathcal{S} \mid \mathbf{Y}_T, \mathcal{G}).
\end{equation}

During training, temporal sequences $\mathcal{T} = \{\mathbf{Y}_{T_s}, \ldots, \mathbf{Y}_{T_s+T_{\text{obs}}-1}\}$ are available as privileged information to guide representation learning, where $T_s$ denotes the start of the training observation window and $T_{\text{obs}}$ denotes the number of observed snapshots, with $T_s+T_{\text{obs}}-1\leq T$. At inference, source prediction uses only the graph $\mathcal{G}$ and the observed snapshot $\mathbf{Y}_T$. This difference in available temporal information creates a training-inference observation mismatch between dynamics-enriched training representations and snapshot-based inference representations.

\subsection{Overview of TGLR}\label{sec4.2}

\begin{figure}[pos=tbp]
    \centering
    \includegraphics[width=\textwidth, trim={1mm 1mm 1mm 0.5mm}, clip]{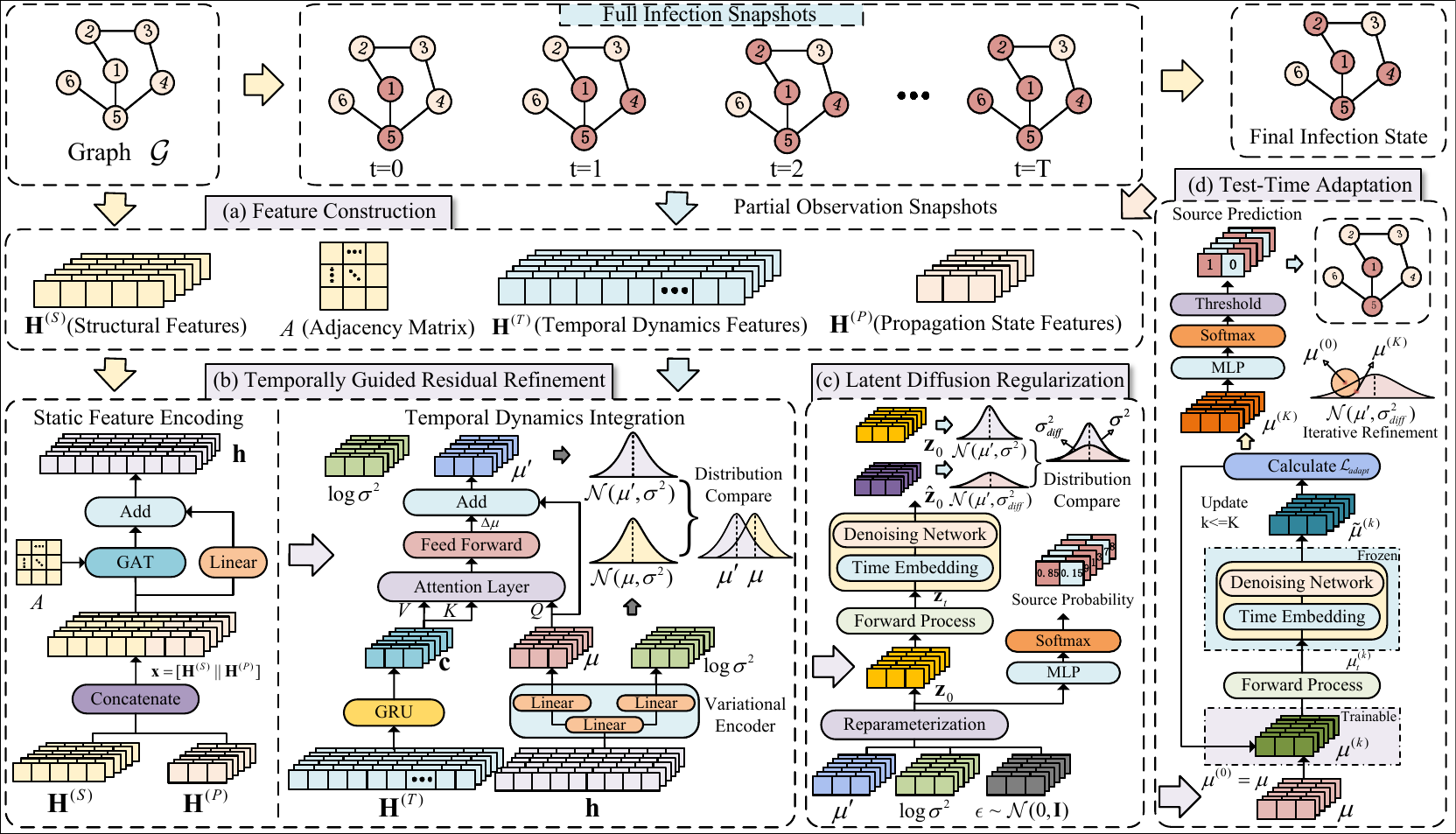}     \caption{Overview of the TGLR architecture. \textbf{(a) Feature Construction} extracts features from graph observations. \textbf{(b) Temporally Guided Residual Refinement} encodes static features into latent distributions and uses GRU-derived temporal dynamics to guide residual refinement of the resulting distributions. \textbf{(c) Latent Diffusion Regularization} broadens and learns distributional structure via denoising networks. \textbf{(d) Test-Time Adaptation} adapts representations for source prediction during inference.}
    \label{fig:method}
\end{figure}

To address the training-inference observation mismatch, TGLR combines temporally guided residual refinement with latent diffusion modeling to learn from temporal propagation sequences while supporting source localization from a single snapshot at inference. The method consists of four stages (Figure~\ref{fig:method}): (a) \textbf{Feature Construction} extracts structural and propagation state features from the observed graph and snapshot, together with temporal dynamics features used only during training; (b) \textbf{Temporally Guided Residual Refinement} encodes snapshot-based features into base latent distributions and uses temporal dynamics to refine their means during training; (c) \textbf{Latent Diffusion Regularization} broadens the refined latent distribution and learns its structure, providing a
distributional prior; and (d) \textbf{Test-Time Adaptation} uses this prior to adapt snapshot-based latent representations toward the dynamics-enriched distribution when temporal information is unavailable during inference.

\subsection{Feature Construction}\label{sec4.3}

To characterize source nodes under the single-snapshot inference setting, we construct node features by integrating information from three perspectives: graph structure, propagation states, and diffusion dynamics. Purely structural features capture the relative position of a node within the network topology but are insufficient to distinguish potential sources when local structures are similar. Propagation state features provide direct cues derived from the observed snapshot. Temporal dynamics features describe statistical changes in neighborhood states within the observation window and are used only during training. Accordingly, three categories of node features are considered to support discriminative source localization: \textbf{structural features}, \textbf{propagation state features}, and \textbf{temporal dynamics features}.

\subsubsection{\textbf{Structural Features}} 
Node roles are characterized from a topological perspective using a set of five classical graph descriptors. 
For each node $v_i$, the structural feature vector $\mathbf{H}^{(S)}(v_i) \in \mathbb{R}^5$ is defined as:
\begin{equation}
    \mathbf{H}^{(S)}(v_i) = \left[ \text{Deg}(v_i), \text{Avg}(v_i), \text{LCC}(v_i), \text{MaxND}(v_i), \text{Core}(v_i) \right],
\end{equation}
which correspond to degree centrality, average neighbor degree, local clustering coefficient, maximum neighbor degree, and k-core number, respectively. These metrics characterize the node's topological profile from several perspectives: degree and neighbor degree reflect local connectivity and neighborhood influence; the clustering coefficient measures local aggregation; maximum neighbor degree captures the connectivity of its most connected neighbor; and the core number reveals the node's hierarchical position within the network core. To ensure numerical stability, features with unbounded value ranges are max-normalized to the interval $[0,1]$ prior to concatenation.

\subsubsection{\textbf{Propagation State Features}} Since static structural metrics cannot distinguish between different cascades on the same graph, the observed snapshot $\mathbf{Y}_T$ serves as the basis for deriving discriminative cues. Accordingly, the propagation state feature vector $\mathbf{H}^{(P)}(v_i) \in \mathbb{R}^3$ is constructed by combining local neighborhood statistics with a global propagation score.

The local infection density of node $v_i$ is defined according to the infection status of its immediate neighborhood as follows:
\begin{equation}
\begin{aligned}
    h_{inf}(v_i) &= \frac{\sum_{u \in \mathcal{N}(v_i)} Y_{T,u}}{|\mathcal{N}(v_i)|}, \\
    h_{uninf}(v_i) &= 1 - h_{inf}(v_i),
\end{aligned}
\end{equation}
where $h_{inf}$ represents the local infection density, and $h_{uninf}$ reflects the local uninfected proportion.

In addition to local neighborhood statistics, global diffusion patterns induced by the observed snapshot are captured through label propagation over the graph structure~\cite{wang2017LPSI}. Starting from the snapshot infection states $\mathbf{Y}_T$, an iterative propagation process is applied to propagate label information across the graph $\mathcal{G}$:
\begin{equation}
    \mathbf{G}_{t+1} = \alpha \mathbf{S} \mathbf{G}_t + (1 - \alpha)\mathbf{Y}_T,
    \label{eq:lpsi_iter}
\end{equation}
where $\mathbf{S} = D^{-1/2}AD^{-1/2}$ is the symmetrically normalized adjacency matrix, and $\alpha \in (0,1)$ controls the fraction of label information propagated from neighboring nodes. Here, $D$ denotes the degree matrix of graph $\mathcal{G}$. The iteration is initialized with $\mathbf{G}_0 = \mathbf{Y}_T$ and is performed for a fixed number of steps. The resulting propagation value at node $v_i$ is denoted as $h_{label}(v_i)$.

Consequently, the final propagation state feature vector integrates local neighborhood statistics and global diffusion information derived from the observed snapshot:
\begin{equation}
    \mathbf{H}^{(P)}(v_i) = \left[ h_{inf}(v_i), h_{uninf}(v_i), h_{label}(v_i) \right] \in \mathbb{R}^3.
    \label{eq:prop_features}
\end{equation}

\subsubsection{\textbf{Temporal Dynamics Features}}
These features are used only during training. They incorporate temporal information to guide representation learning and are constructed from infection snapshots observed over an observation window of length $T_{\text{obs}}$. 

For each node $v_i$, four temporal statistics are defined to characterize the evolution of neighborhood propagation states over the observation window. The transition rates are computed for $t=T_s+1,\dots,T_s+T_{\text{obs}}-1$, while the neighborhood infected and uninfected proportions are computed for $t=T_s,\dots,T_s+T_{\text{obs}}-1$. First, the infection rate among uninfected neighbors quantifies the proportion of newly infected neighbors between adjacent observations and is defined as
\begin{equation}
    \lambda^{(t)}(v_i) = 
    \frac{\sum_{u \in \mathcal{N}(v_i)} \mathbb{I}\!\left(Y_{t,u}=1 \wedge Y_{t-1,u}=0\right)}
    {\sum_{u \in \mathcal{N}(v_i)} \mathbb{I}\!\left(Y_{t-1,u}=0\right)}.
\end{equation}
Second, the recovery rate among infected neighbors is defined analogously as
\begin{equation}
    \gamma^{(t)}(v_i) =
    \frac{\sum_{u \in \mathcal{N}(v_i)} \mathbb{I}\!\left(Y_{t,u}=0 \wedge Y_{t-1,u}=1\right)}
    {\sum_{u \in \mathcal{N}(v_i)} \mathbb{I}\!\left(Y_{t-1,u}=1\right)},
\end{equation}
for monotonic diffusion processes, such as the SI model and real-world information propagation cascades, infected states remain persistent once acquired, resulting in $\gamma^{(t)}(v_i)=0$. This definition enables the temporal representation to support diffusion processes with and without recovery dynamics.

In addition to transition rates, the proportion of infected neighbors at each observation time is defined as
\begin{equation}
    \rho_I^{(t)}(v_i) =
    \frac{\sum_{u \in \mathcal{N}(v_i)} Y_{t,u}}{|\mathcal{N}(v_i)|},
\end{equation}
similarly, the proportion of uninfected neighbors is given by
\begin{equation}
    \rho_S^{(t)}(v_i) =
    \frac{\sum_{u \in \mathcal{N}(v_i)} (1 - Y_{t,u})}{|\mathcal{N}(v_i)|}.
\end{equation}

The temporal dimensions of these statistics follow their definitions: $\boldsymbol{\lambda}(v_i)$ and $\boldsymbol{\gamma}(v_i)$ contain $T_{\text{obs}}-1$ transition statistics, while $\boldsymbol{\rho}_I(v_i)$ and $\boldsymbol{\rho}_S(v_i)$ contain $T_{\text{obs}}$ neighborhood state proportions. Finally, the temporal dynamics feature vector for node $v_i$ is obtained by concatenating these temporal statistics:
\begin{equation}
    \mathbf{H}^{(T)}(v_i)
    =
    \big[
    \boldsymbol{\lambda}(v_i),~
    \boldsymbol{\gamma}(v_i),~
    \boldsymbol{\rho}_I(v_i),~
    \boldsymbol{\rho}_S(v_i)
    \big] \in \mathbb{R}^{4T_{\text{obs}}-2}.
\end{equation}

\subsection{Temporally Guided Residual Refinement}\label{sec4.4}

TGLR mitigates the training-inference observation mismatch through progressive latent refinement. During training, temporal propagation dynamics are encoded as node-level temporal contexts and used to guide residual refinement of latent distributions as follows:

\subsubsection{\textbf{Static Feature Encoding}}

Given the concatenated vector of static features 
$\mathbf{x}_i = [\mathbf{H}^{(S)}(v_i) \parallel \mathbf{H}^{(P)}(v_i)] \in \mathbb{R}^{8}$ 
for all nodes $v_i \in \mathcal{V}$, we employ a GNN backbone to integrate information from neighboring nodes. Formally, the embedding $\mathbf{h}_i$ for node $v_i$ is computed via a message-passing function $f_{\text{agg}}$ with residual connections~\cite{he2016deep, gat17}:

\begin{equation}
\mathbf{h}_i = f_{\text{agg}}\left(\mathbf{x}_i, \{\mathbf{x}_j : v_j \in \mathcal{N}_i\}\right) + f_{\text{res}}(\mathbf{x}_i) \in \mathbb{R}^{d_h},
\label{eq:gnn_agg}
\end{equation}
where $\mathcal{N}_i = \{v_j : A_{ij} = 1\}$ denotes the set of neighbors of node 
$v_i$ as specified by the adjacency matrix $A$, and $f_{\text{res}}$ is a learnable 
projection for dimension matching. Representative architectures such as Graph Convolutional Networks (GCN)~\cite{kipf2016gcn} and Graph Sample and Aggregate (GraphSAGE)~\cite{hamilton2017graphsage} can be used to instantiate $f_{\mathrm{agg}}$. In our implementation, we adopt the Graph Attention 
Network (GAT)~\cite{gat17}, which provides a flexible mechanism for neighborhood 
aggregation.

\subsubsection{\textbf{Temporal Dynamics Integration}}
\label{sec:tglr}

The embedding $\mathbf{h}_i$ for each node $v_i \in \mathcal{V}$ is encoded via learnable networks $\phi_\mu, \phi_\sigma : \mathbb{R}^{d_h} \rightarrow \mathbb{R}^{d_z}$ into a probabilistic latent representation $q(\mathbf{z}_i | \mathbf{x}_i) = \mathcal{N}(\boldsymbol{\mu}_i, \text{diag}(\boldsymbol{\sigma}_i^2))$:
\begin{equation}
\boldsymbol{\mu}_i = \phi_{\mu}(\mathbf{h}_i), \quad 
\log \boldsymbol{\sigma}_i^2 = \phi_{\sigma}(\mathbf{h}_i).
\end{equation}

During training, when temporal propagation sequences are available, temporal dynamics guide residual refinement of the latent distribution. Specifically, we leverage temporal features $\mathbf{H}^{(T)}(v_i) \in \mathbb{R}^{4T_{\text{obs}}-2}$ (Section~\ref{sec4.3}) to capture neighborhood infection dynamics over the observation window. A GRU processes these features and extracts the final hidden state as an aggregated temporal representation:
\begin{equation}
\mathbf{c}_i = \mathrm{GRU}(\mathbf{H}^{(T)}(v_i)) \in \mathbb{R}^{d_z},
\label{eq:temporal_context}
\end{equation}
where $\mathbf{c}_i$ summarizes the temporal propagation patterns relevant to node $v_i$.

To incorporate temporal dynamics into the latent distribution, the static latent mean $\boldsymbol{\mu}_i$ and temporal context $\mathbf{c}_i$ are fed into an attention layer followed by an FFN to estimate a residual refinement term for the distribution mean:
\begin{equation}
\Delta\boldsymbol{\mu}_i
=
g\left(
\mathcal{A}(\boldsymbol{\mu}_i, \mathbf{c}_i)
\right),
\label{eq:delta_mu}
\end{equation}
where $\mathcal{A}(\cdot)$ denotes a learnable attention layer that uses $\boldsymbol{\mu}_i$ as the query and $\mathbf{c}_i$ as the key and value context, and $g(\cdot)$ denotes a learnable FFN. The refined mean is obtained through a residual connection:
\begin{equation}
\boldsymbol{\mu}_i' = \boldsymbol{\mu}_i + \Delta\boldsymbol{\mu}_i.
\label{eq:refined_mean}
\end{equation}

\textbf{Latent-level residual refinement design}. We adopt a latent-level residual refinement strategy that contrasts with feature-level concatenation approaches. While concatenation directly fuses static and temporal features before encoding, our approach maintains an explicit base distribution $\boldsymbol{\mu}_i$ for static information and applies temporal refinements $\Delta\boldsymbol{\mu}_i$ in the latent space. 

This design enables distinct encoder behaviors during training and inference. During training, the refined distribution 
$q(\mathbf{z}_i | \mathbf{x}_i, \mathbf{H}^{(T)}(v_i)) = \mathcal{N}(\boldsymbol{\mu}_i', \boldsymbol{\sigma}_i^2)$ 
incorporates temporal information through the refined mean $\boldsymbol{\mu}_i'$. At inference, the encoder produces 
$q(\mathbf{z}_i | \mathbf{x}_i) = \mathcal{N}(\boldsymbol{\mu}_i, \boldsymbol{\sigma}_i^2)$ from static features alone. The probabilistic formulation provides a key advantage: by sampling from the distribution around $\boldsymbol{\mu}_i'$, the model learns 
to handle a neighborhood of latent representations defined by $\boldsymbol{\sigma}_i^2$. This learned tolerance enables robust 
predictions even when $\boldsymbol{\mu}_i \neq \boldsymbol{\mu}_i'$, provided $\boldsymbol{\mu}_i$ falls within the training neighborhood. Design Choice Validation (Section~\ref{sec:temproal_integration}) validates this design, showing better performance than the concatenation strategy in most settings.

\subsection{Latent Diffusion Regularization}

To broaden and learn the structure of latent distributions for robust representation and adaptation, we employ a diffusion probabilistic model on latent representations. Operating in the latent space, similar to Graphusion~\cite{graphusion24}, allows for efficient modeling of distributional structure while avoiding the high dimensionality of raw graph space.

Given $\mathbf{z}_0 \sim \mathcal{N}(\boldsymbol{\mu}_i', \boldsymbol{\sigma}_i^2)$ sampled from the encoder output, the forward process corrupts the latent with Gaussian noise:
\begin{equation}
    q(\mathbf{z}_t | \mathbf{z}_0) = \mathcal{N}(\mathbf{z}_t; \sqrt{\bar{\alpha}_t}\mathbf{z}_0, (1-\bar{\alpha}_t)I),
    \label{eq:forward_diff}
\end{equation}
where $\bar{\alpha}_t = \prod_{s=1}^{t} (1-\beta_s)$ with noise schedule $\{\beta_t\}_{t=1}^{T_{\text{diff}}}$. This enables direct sampling without iterative forward steps:
\begin{equation}
    \mathbf{z}_t = \sqrt{\bar{\alpha}_t}\mathbf{z}_0 + \sqrt{1-\bar{\alpha}_t}\boldsymbol{\epsilon}, \quad \boldsymbol{\epsilon} \sim \mathcal{N}(\mathbf{0}, I).
    \label{eq:sampling}
\end{equation}

We train a neural network $f_\theta$ to directly predict the clean 
latent $\mathbf{z}_0$ from the noisy state $\mathbf{z}_t$:
\begin{equation}
    \hat{\mathbf{z}}_0 = f_\theta(\mathbf{z}_t, t).
    \label{eq:direct_prediction}
\end{equation}

While DDPM~\cite{ddpm20} discusses both $\boldsymbol{\epsilon}$ and $\mathbf{z}_0$ parameterizations, we adopt the latter as it allows 
$f_\theta$ to serve as an explicit distributional prior over latent representations, directly consumable during test-time adaptation 
without additional conversion. The denoising distribution is then modeled as:
\begin{equation}
    p_\theta(\mathbf{z}_0 | \mathbf{z}_t) = \mathcal{N}\!\left(\mathbf{z}_0;\, f_\theta(\mathbf{z}_t, t),\, \sigma^2 I\right).
    \label{eq:denoising_dist}
\end{equation}

This design serves two purposes: (1) \textit{Inference efficiency}: During inference, $f_\theta$ directly predicts $\hat{\mathbf{z}}_0$ from a noisy latent sample, replacing the $T$ sequential denoising steps in standard DDPM sampling~\cite{ddpm20}. (2) \textit{Distributional regularization}: By learning to reconstruct clean latent representations from noise-corrupted versions across varying noise levels, 
$f_\theta$ broadens the latent distribution, shaping the encoder output toward a wider distributional space with increased variance $\sigma_{\text{diff}}^2 > \sigma^2$, as illustrated in Figure~\ref{fig:method}. The theoretical intuition and quantitative validation of this broadening effect are provided in Appendix~\ref{app:variance}. The trained $f_\theta$ is subsequently employed as a distributional prior during test-time adaptation, as described in Section~\ref{sec:test_time}.

\subsection{Training Objective}
\label{sec:training_obj}

Our goal is to maximize $p(\mathcal{S}|\mathbf{Y}_T, \mathcal{G})$ at inference, where only the final snapshot $\mathbf{Y}_T$ is available. During training, we leverage privileged temporal information $\mathcal{T}$ through refinement mechanisms (Section~\ref{sec:tglr}) to learn a dynamics-enriched latent space that generalizes to the static final observation at inference. We aim to maximize the following conditional log-likelihood during training:
\begin{equation}
\log p(\mathcal{S}|\mathbf{Y}_T, \mathcal{G}, \mathcal{T}).
\label{eq:training_objective}
\end{equation}

\subsubsection{\textbf{Discriminative variational inference}}

We introduce latent variables $\mathbf{z}$ to model uncertainty relevant to 
source localization. The joint distribution decomposes as:
\begin{equation}
\log p(\mathbf{Y}_T, \mathcal{G}, \mathcal{T}, \mathcal{S}) 
= \log p(\mathcal{S}|\mathbf{Y}_T, \mathcal{G}, \mathcal{T}) 
+ \log p(\mathbf{Y}_T, \mathcal{G}, \mathcal{T}).
\label{eq:joint_decomp}
\end{equation}

Introducing a variational distribution 
$q(\mathbf{z}|\mathbf{Y}_T, \mathcal{G}, \mathcal{T})$ and applying 
Jensen's inequality:
\begin{align}
&\log p(\mathbf{Y}_T, \mathcal{G}, \mathcal{T}, \mathcal{S}) \nonumber \\
&= \log \int p(\mathbf{Y}_T, \mathcal{G}, \mathcal{T}, \mathcal{S}, \mathbf{z}) d\mathbf{z} \nonumber \\
&= \log \int \frac{q(\mathbf{z}|\mathbf{Y}_T, \mathcal{G}, \mathcal{T})}{q(\mathbf{z}|\mathbf{Y}_T, \mathcal{G}, \mathcal{T})} 
p(\mathbf{Y}_T, \mathcal{G}, \mathcal{T}, \mathcal{S}, \mathbf{z}) d\mathbf{z} \nonumber \\
&\geq \mathbb{E}_{q(\mathbf{z}|\mathbf{Y}_T,\mathcal{G},\mathcal{T})}\left[
\log \frac{p(\mathbf{Y}_T, \mathcal{G}, \mathcal{T}, \mathcal{S}, \mathbf{z})}{q(\mathbf{z}|\mathbf{Y}_T, \mathcal{G}, \mathcal{T})}
\right] \nonumber \\
&= \mathbb{E}_{q}[\log p(\mathbf{Y}_T, \mathcal{G}, \mathcal{T}|\mathbf{z})] 
+ \mathbb{E}_{q}[\log p(\mathcal{S}|\mathbf{z})] \nonumber \\
&\quad + \mathbb{E}_{q}[\log p(\mathbf{z})] 
- \mathbb{E}_{q}[\log q(\mathbf{z}|\mathbf{Y}_T, \mathcal{G}, \mathcal{T})] \nonumber \\
&= \mathbb{E}_{q}[\log p(\mathbf{Y}_T, \mathcal{G}, \mathcal{T}|\mathbf{z})] 
+ \mathbb{E}_{q}[\log p(\mathcal{S}|\mathbf{z})] \nonumber \\
&\quad - \text{KL}(q(\mathbf{z}|\mathbf{Y}_T,\mathcal{G},\mathcal{T}) \| p(\mathbf{z})).
\label{eq:elbo_joint}
\end{align}

Combining with Eq.~\eqref{eq:joint_decomp}:
\begin{align}
&\log p(\mathcal{S}|\mathbf{Y}_T, \mathcal{G}, \mathcal{T}) \nonumber \\
&= \log p(\mathbf{Y}_T, \mathcal{G}, \mathcal{T}, \mathcal{S}) 
- \log p(\mathbf{Y}_T, \mathcal{G}, \mathcal{T}) \nonumber \\
&\geq \mathbb{E}_{q}[\log p(\mathbf{Y}_T, \mathcal{G}, \mathcal{T}|\mathbf{z})] 
+ \mathbb{E}_{q}[\log p(\mathcal{S}|\mathbf{z})] \nonumber \\
&\quad - \text{KL}(q(\mathbf{z}|\mathbf{Y}_T,\mathcal{G},\mathcal{T}) \| p(\mathbf{z})) 
- \log p(\mathbf{Y}_T, \mathcal{G}, \mathcal{T}) \nonumber \\
&= \mathbb{E}_{q}[\log p(\mathbf{Y}_T, \mathcal{G}, \mathcal{T}|\mathbf{z}) 
- \log p(\mathbf{Y}_T, \mathcal{G}, \mathcal{T})] \nonumber \\
&\quad + \mathbb{E}_{q}[\log p(\mathcal{S}|\mathbf{z})] 
- \text{KL}(q(\mathbf{z}|\mathbf{Y}_T,\mathcal{G},\mathcal{T}) \| p(\mathbf{z})).
\label{eq:conditional_bound}
\end{align}

The first term measures how much latent variables improve observation likelihood. For discriminative prediction focused on source identification 
rather than observation reconstruction, we assume 
$p(\mathbf{Y}_T, \mathcal{G}, \mathcal{T}|\mathbf{z}) \approx p(\mathbf{Y}_T, \mathcal{G}, \mathcal{T})$, 
yielding:
\begin{align}
\log p(\mathcal{S}|\mathbf{Y}_T, \mathcal{G}, \mathcal{T}) 
&\gtrsim \mathbb{E}_{q(\mathbf{z}|\mathbf{Y}_T,\mathcal{G},\mathcal{T})}[\log p(\mathcal{S}|\mathbf{z})] \nonumber \\
&\quad - \text{KL}(q(\mathbf{z}|\mathbf{Y}_T,\mathcal{G},\mathcal{T}) \| p(\mathbf{z})).
\label{eq:discriminative_obj}
\end{align}

To optimize this objective, we minimize:
\begin{align}
\mathcal{L}_{\text{base}} &= \mathcal{L}_{\text{pred}} + \lambda_{\text{KL}} \cdot \mathcal{L}_{\text{KL}} \nonumber \\
&= -\mathbb{E}_{q(\mathbf{z}|\mathbf{x}, \mathbf{H}^{(T)})}[\log p(\mathcal{S}|\mathbf{z})] \nonumber \\
&\quad + \lambda_{\text{KL}} \cdot \text{KL}(q(\mathbf{z}|\mathbf{x}, \mathbf{H}^{(T)}) \| p(\mathbf{z})),
\label{eq:base_loss}
\end{align}
where the encoder $q(\mathbf{z}|\mathbf{x}, \mathbf{H}^{(T)})$ uses 
temporally guided residual refinement to update the latent means (Section~\ref{sec:tglr}), $p(\mathbf{z}) = \mathcal{N}(\mathbf{0}, I)$ is the prior, 
$\mathcal{L}_{\text{pred}}$ is the cross-entropy between predicted source 
probabilities and ground-truth labels, and $\mathcal{L}_{\text{KL}}$ is 
computed in closed form against the standard Gaussian prior:
\begin{equation}
\mathcal{L}_{\text{KL}} = \frac{1}{2n}\sum_{i=1}^{n}
\sum_{j=1}^{d_z}
\left((\mu'_{i,j})^2 + \sigma_{i,j}^2 - \log\sigma_{i,j}^2 - 1\right),
\label{eq:kl_loss}
\end{equation}
where $\mu'_{i,j}$ and $\sigma_{i,j}^2$ are the $j$-th components of the temporally refined mean and the unchanged variance for node $i$. This objective resembles the Variational Information Bottleneck (VIB)~\cite{alemi2016deep} in a discriminative graph setting, where $\lambda_{\text{KL}}$ plays the role of the bottleneck coefficient $\beta$ balancing compression and prediction. Unlike standard VIB on static inputs, our encoder parameterizes $q(\mathbf{z}|\mathbf{x}, \mathbf{H}^{(T)})$ during training, encouraging the latent space to capture dynamics-relevant structure unavailable at inference.

\subsubsection{\textbf{Latent Diffusion Regularization}}

Beyond the base variational objective, we introduce diffusion-based regularization to learn the structure of the latent distribution. Given refined latent samples $\mathbf{z}_0 \sim q(\mathbf{z}|\mathbf{x}, \mathbf{H}^{(T)})$, noisy latent samples $\mathbf{z}_t$ are obtained according to Eq.~\eqref{eq:sampling}. The diffusion model is then trained to reconstruct clean latent representations:
\begin{equation}
\mathcal{L}_{\text{diff}}
=
\mathbb{E}_{t,\mathbf z_0,\boldsymbol\epsilon}
\left[
\left\|
\mathbf z_0-
f_\theta(\mathbf z_t,t)
\right\|^2
\right],
\label{eq:diff_loss}
\end{equation}
where the reconstruction target $\mathbf{z}_0$ is detached to stabilize training while allowing gradients to flow through $f_\theta$.

\subsubsection{\textbf{Complete Training Objective}}

The complete objective is:
\begin{equation}
\mathcal{L}_{\text{total}} = \mathcal{L}_{\text{pred}} + \lambda_{\text{KL}} \cdot 
\mathcal{L}_{\text{KL}} + \lambda_{\text{diff}} \cdot \mathcal{L}_{\text{diff}},
\label{eq:total_loss}
\end{equation}
where $\lambda_{\text{KL}}$ controls the strength of KL regularization and 
$\lambda_{\text{diff}}$ balances the diffusion regularization. This joint 
training ensures the learned latent space is both discriminative for source 
prediction and broadly covered for robust test-time adaptation.

\subsection{Test-Time Adaptation}

\label{sec:test_time}

At inference, temporal sequences are unavailable, leaving only the final snapshot $\mathbf{Y}_T$ and graph $\mathcal{G}$. The encoder produces 
base means $\boldsymbol{\mu}_i$ without temporal refinement. We leverage the trained diffusion model $f_\theta$ as a distributional prior to adapt 
these means toward the dynamics-enriched latent distribution learned during training. Given the static mean $\boldsymbol{\mu}_i$, we perform $K$ 
gradient-based refinement iterations:
\begin{equation}
\boldsymbol{\mu}_i^{(k+1)} = \boldsymbol{\mu}_i^{(k)} - \eta 
\nabla_{\boldsymbol{\mu}_i^{(k)}} \mathcal{L}_{\text{adapt}},
\label{eq:tta_update}
\end{equation}
where the adaptation loss uses the diffusion model as a distributional prior:
\begin{equation}
\mathcal{L}_{\text{adapt}} = \|\tilde{\boldsymbol{\mu}}_i^{(k)} - 
\boldsymbol{\mu}_i^{(k)}\|^2,
\label{eq:tta_loss}
\end{equation}
where $\tilde{\boldsymbol{\mu}}_i^{(k)}=f_\theta(\boldsymbol{\mu}_{i,t}^{(k)},t)$ denotes the denoised prediction and $\boldsymbol{\mu}_{i,t}^{(k)} = \sqrt{\bar{\alpha}_t}\boldsymbol{\mu}_i^{(k)} + \sqrt{1-\bar{\alpha}_t}\boldsymbol{\epsilon}$ is the noisy latent at iteration $k$, with $t$ uniformly sampled from $\{1,\ldots,T_{\text{diff}}\}$ per node per iteration. By minimizing this reconstruction error, 
$\boldsymbol{\mu}_i$ is iteratively guided toward the broader latent distribution captured by $f_\theta$, bridging the gap between static 
observations and the dynamics-enriched representations learned during training. The parameters of both the encoder and $f_\theta$ remain frozen 
throughout adaptation; only the latent means $\{\boldsymbol{\mu}_i^{(k)}\}$ are updated. After $K$ iterations, the refined means 
$\{\boldsymbol{\mu}_i^{(K)}\}$ are used for prediction.

This design departs from standard diffusion inference in two respects. First, since our goal is representation refinement rather than generation, we use $f_\theta$ as a single-step corrector, avoiding the $\mathcal{O}(T_{\text{diff}})$ overhead of full iterative denoising. Second, the noising step maintains consistency with the training procedure. Furthermore, by sampling different noise levels $t$ and realizations $\boldsymbol{\epsilon}$ across iterations, the gradient signals probe the broadened latent distribution from multiple directions, which is particularly beneficial for nodes at the boundary of or outside the training distribution, gradually steering their representations toward the dynamics-enriched latent space.

\section{Experiments}

We conduct comprehensive experiments to evaluate TGLR across multiple dimensions. Specifically, our evaluation is guided by the following research questions:

\begin{itemize}
    \item \textbf{RQ1 (Effectiveness):} How does TGLR compare with state-of-the-art source localization methods across different network structures and diffusion scenarios?

    \item \textbf{RQ2 (Component Contribution):} What is the individual contribution of each component---temporally guided residual refinement, latent diffusion regularization, and test-time adaptation?

    \item \textbf{RQ3 (Design Validation):} Are the key design choices empirically justified: (i) latent-level residual refinement versus feature-level concatenation for temporal integration, and (ii) broadening the latent distribution to accommodate training-inference variability versus reducing the gap through explicit distribution alignment?

    \item \textbf{RQ4 (Hyperparameter Sensitivity):} How do training hyperparameters affect model performance?

    \item \textbf{RQ5 (Visualization):} Can we visualize the training-inference distributional shift and how TGLR mitigates it through distributional bridging mechanisms?

    \item \textbf{RQ6 (Efficiency):} What is the computational cost of TGLR in terms of training and inference time compared to baseline methods?
\end{itemize}

\subsection{Datasets}

We conduct experiments on eight datasets: four simulated propagation datasets constructed on real-world network structures and four real-world propagation datasets collected from social media platforms. This combination allows us to evaluate model performance under both simulated and realistic propagation scenarios. For fair comparison, all methods are evaluated under the same dataset configurations.

\subsubsection{Simulated Propagation Datasets}

\begin{table}[pos=tbp]
\centering
\caption{Statistical properties of simulated propagation datasets.}
\label{tab:simulated_stats}
\csffittable{
\begin{tabular}{lcccc}
\toprule
\textbf{Property} & \textbf{Karate} & \textbf{Jazz} & \textbf{Cora-ML} & \textbf{Facebook} \\
\midrule
Nodes              & 34     & 198    & 2,810  & 4,039   \\
Edges              & 78     & 2,742  & 7,981  & 88,234  \\
Avg Degree         & 4.588   & 27.697  & 5.680   & 43.691   \\
Clustering         & 0.571   & 0.617   & 0.278   & 0.605    \\
Density            & 0.139  & 0.141  & 0.002  & 0.011   \\
Diameter           & 5      & 6      & 17     & 8       \\
\bottomrule
\end{tabular}
}
\end{table}

To systematically evaluate source localization under controlled propagation dynamics across diverse network topologies, we simulate information diffusion on four representative real-world networks: \textbf{Karate}~\cite{karate}, \textbf{Jazz}~\cite{jazz}, \textbf{Cora-ML}~\cite{cora_ml}, and \textbf{Facebook}~\cite{facebook}. 

Following previous source localization studies~\cite{chen22slvae,liu2024DDSL}, we simulate propagation cascades using both the SI and SIR models, with 10\% of nodes randomly selected as ground-truth sources in each simulation run. These models allow evaluation under propagation dynamics without and with recovery, respectively. We continue the diffusion process until 30\% of the network is infected. We repeat this process 100 times for each network under each propagation model, generating 100 cascades per dataset. Additional simulation parameters are provided in Table~\ref{tab:simulation_params} in the Appendix. Table~\ref{tab:simulated_stats} summarizes the structural properties of these networks.

\subsubsection{Real-World Propagation Datasets}

In addition to simulated propagation datasets, we evaluate TGLR on three widely used real-world datasets and our newly constructed Twitter25 dataset. Among them, Weibo includes reposts, replies, and quoting interactions~\cite{ma2016rumors}, whereas Twitter15 and Twitter16 only contain retweet and reply cascades~\cite{ma2017propagation}. However, real-world information propagation involves diverse user interactions. Quote tweets provide an additional channel for information diffusion by allowing users to redistribute existing tweets with their own commentary~\cite{garimella2016quote,flamino2023polarization,tokita2024measuring}. To capture this additional diffusion mechanism, we construct Twitter25 by incorporating quoting interactions, consistent with their inclusion in the Weibo dataset. The four datasets are described as follows:

\begin{figure}[pos=tbp]
    \centering
    \makebox[\linewidth][c]{\hspace*{0.118783\columnwidth}%
      \includegraphics[width=0.519036\columnwidth]{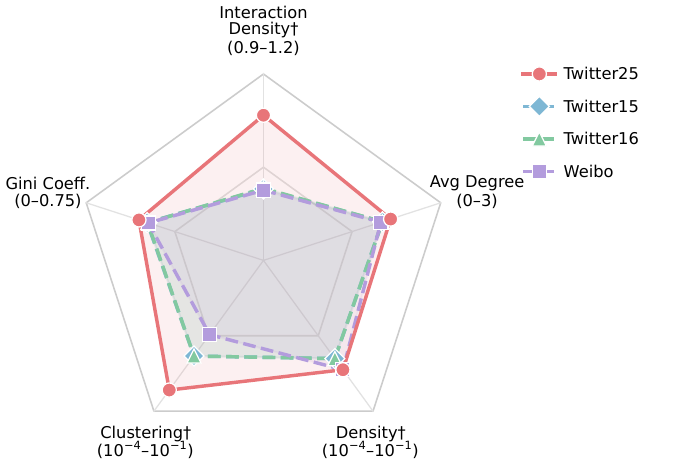}}
    \caption{
       Structural comparison of four propagation datasets across five dimensions: Interaction Density, Average Degree, Graph Density, Clustering Coefficient, and Gini Coefficient. Axes marked with~$\dagger$ are normalized on a logarithmic scale; remaining axes use linear normalization. Each axis is normalized to the displayed range. Numerical values before normalization are provided in Table~\ref{tab:realworld_structural_stats} in Appendix~\ref{app:realworld_stats}.
    }
    \label{fig:radar}
\end{figure}

\begin{table}[pos=tbp]
\centering
\caption{Sizes and interaction types of real-world propagation datasets.}
\label{tab:dataset_stats}
\setlength{\tabcolsep}{4pt}
\csffittable{\begin{tabular}{lcccc}
\toprule
 & \textbf{Twitter25} & \textbf{Twitter15} & \textbf{Twitter16} & \textbf{Weibo} \\
\midrule
\multicolumn{5}{l}{\textit{Size statistics}} \\
Users       & 564,335 & 599,196 & 347,828 & 3,696,999 \\
Edges       & 635,183 & 602,105 & 349,958 & 3,706,477 \\
Cascades    & 981     & 1,490   & 818     & 4,664     \\
\midrule
\multicolumn{5}{l}{\textit{Interaction types}} \\
Retweets   & \cmark & \cmark & \cmark & \cmark \\
Replies              & \cmark & \cmark & \cmark & \cmark \\
Quotes         & \cmark & \xmark & \xmark & \cmark \\
\bottomrule
\end{tabular}}
\end{table}

\begin{itemize}
    \item \textbf{Twitter15}~\cite{ma2017propagation}: This dataset contains tweet cascades formed by retweets and replies collected from the Twitter platform in 2015. Each cascade corresponds to a single propagation instance, accompanied by manually annotated labels indicating whether the content is a rumor or non-rumor.
    
    \item \textbf{Twitter16}~\cite{ma2017propagation}: Similar in structure to Twitter15, this dataset comprises cascades from the year 2016, with the same rumor annotation schema. Both Twitter15 and Twitter16 are widely used in rumor classification and propagation analysis tasks.
    
    \item \textbf{Weibo}~\cite{ma2016rumors}: This dataset contains propagation cascades for 4,664 events collected from Sina Weibo. Each event includes an original post and its reposts and replies, together with an event-level rumor or non-rumor label.

    \item \textbf{Twitter25}: This dataset contains propagation cascades from Twitter (now X) associated with events occurring between January and June 2025. It covers retweets, quotes, and replies.
\end{itemize}

It is worth mentioning that we evaluate TGLR and all baseline methods on Twitter25 both with and without quoting interactions. In the setting without quote tweets, denoted as Twitter25-NQ, we use only retweet and reply interactions, aligning the interaction types with those included in Twitter15 and Twitter16. In the setting with quote tweets, we keep all three interaction types, consistent with Weibo, to assess source localization performance under the training--inference observation mismatch.

To construct Twitter25, we verified event veracity using reputable fact-checking organizations, including Snopes\footnote{\url{https://www.snopes.com/fact-check/}} and PolitiFact\footnote{\url{https://www.politifact.com/factchecks/2025/}}. For each collected source tweet, we used the official API to collect these interactions and preserve the relationships between quotes and the posts they quote, as well as between replies and the posts they respond to.

Table~\ref{tab:dataset_stats} summarizes the sizes and interaction types of the four datasets, while Figure~\ref{fig:radar} compares their observed structural characteristics. Five structural dimensions are considered. Interaction density quantifies the average number of interactions per user aggregated across all cascades. Average degree and graph density describe overall connectivity. The clustering coefficient measures how densely users interact within local neighborhoods, while the Gini coefficient characterizes the inequality of the degree distribution. In the collected graphs, Twitter25 has a higher interaction density and a higher clustering coefficient than the other datasets, indicating more interactions per user and greater local interconnectedness. Its higher Gini coefficient reflects a more unequal degree distribution. Twitter15 and Twitter16 exhibit similar values across all five dimensions, whereas Weibo, despite its larger scale, has the lowest interaction density and clustering coefficient among the four datasets.

Following existing source localization studies on real-world cascades~\cite{SIDSL2025}, ground-truth sources in all real-world datasets are defined as the earliest $10\%$ of users ranked by participation timestamps in each cascade. User profiles and content features are not used by any of the evaluated methods.

\subsection{Baselines}

We compare the proposed TGLR with a diverse set of representative baselines, covering infection-state label propagation, GNNs, and probabilistic generative models:
\begin{itemize}
     \item \textbf{LPSI}~\cite{wang2017LPSI}: A label propagation-based method that iteratively diffuses infection labels through the network to infer the most likely source, without assuming a known propagation model.
    
    \item \textbf{GCNSI}~\cite{ming19GCNSI}: This method incorporates infection features from label propagation into a graph convolutional network, enabling end-to-end learning of multi-hop structural dependencies for source localization.

    \item \textbf{IVGD}~\cite{IVGD22}: An invertible graph diffusion framework that formulates source localization as an inverse problem, using graph residual scenarios for invertibility and validity-aware layers to ensure feasible source predictions.
    
    \item \textbf{SL-VAE}~\cite{chen22slvae}: A variational autoencoder model that encodes infection probabilities into latent variables, explicitly capturing uncertainty in the localization process.

    \item \textbf{DDSL}~\cite{liu2024DDSL}: A denoising diffusion probabilistic model that addresses source uncertainty by generating diverse candidates through a DDPM-based generator and an invertible GNN that incorporates diffusion properties.

    \item \textbf{SIDSL}~\cite{SIDSL2025}: A structure-prior guided diffusion framework leverages graph label propagation to construct topology-aware priors, which are injected into a diffusion-based inference process to improve robustness under limited data.

    \item \textbf{GDFSL}~\cite{GDFSL2025}: A generalized diffusion framework redefines the forward diffusion process to implicitly learn propagation dynamics, enabling source inference across diverse diffusion mechanisms without relying on explicit propagation assumptions.    

    \item \textbf{GTSI}~\cite{lu2025GTSI}: A Graph Transformer framework that captures long-range dependencies through diffusion-aware attention mechanisms, overcoming the local perception limitations of traditional GNNs.

    \item \textbf{PDSL}~\cite{PDSL2026}: Propagation dynamics are modeled in continuous time via Graph Neural ODEs, while a conditional generative model estimates source distributions and refines inference through feedback from simulated propagation behavior.

\end{itemize}

\subsection{Evaluation Metrics}

We evaluate source localization performance using two complementary metrics. In source localization tasks, datasets are typically highly imbalanced with source nodes being the minority class. \textbf{F1 score (F1)}, the harmonic mean of precision and recall, provides a balanced measure particularly important for such imbalanced scenarios, capturing the precision-recall trade-off at a specific decision threshold. \textbf{Area Under the ROC Curve (AUC)} is a threshold-independent metric that measures the model's ranking ability across all possible decision thresholds. Together, F1 evaluates threshold-specific classification performance, while AUC provides a comprehensive assessment of overall discrimination capability.

\subsection{Experimental Setup}

\textbf{Experimental Environment.} 
All experiments are conducted on a server equipped with an NVIDIA RTX A6000 GPU (48GB) running Python 3.10.18. We implement our models using PyTorch 2.5.1. For each dataset, we randomly split the propagation samples into training, validation, and test sets with a ratio of 8:1:1. The validation set is used for hyperparameter tuning.

\textbf{Model Architecture.} TGLR uses a two-layer GAT encoder with 4 attention heads, where the hidden dimension is set to 32 and the output dimension is set to 16. The latent space dimension is set to 32 and the VAE encoder hidden dimension is set to 32. For temporal context modeling, we employ a single-layer GRU where the input dimension is set to 4 and the hidden dimension is set to 32. The residual refinement module uses a lightweight attention layer with two heads, followed by a feed-forward network with an intermediate dimension of 64. The diffusion model employs $T_{\text{diff}}=100$ timesteps with a linear noise schedule from $\beta_{\text{start}} = 1 \times 10^{-4}$ to $\beta_{\text{end}} = 0.01$. The time embedding dimension is set to 16, and the denoising predictor uses a hidden dimension of 32 with 3 residual blocks. The final prediction head is a two-layer MLP where the hidden dimension is set to 32, with ReLU activation and a dropout rate of 0.2. Node features consist of structural properties, propagation states, and temporal propagation dynamics, with dimensions of $5$, $3$, and $4T_{\text{obs}}-2$, respectively. The resulting node feature dimension is therefore $4T_{\text{obs}}+6$ during training.

\textbf{Training and Inference Configuration.}
We train all models using the Adam optimizer with a fixed learning rate of 0.001 for a maximum of 300 epochs. Early stopping is triggered if the validation F1 does not improve for 50 consecutive epochs. The batch size is set to 1, and the loss weights are configured as $\lambda_{\text{KL}} = 0.01$, while $\lambda_{\text{diff}}$ is dataset-dependent. Gradient clipping with a maximum norm of 1.0 is applied to ensure training stability. During inference, we perform test-time adaptation using stochastic gradient descent with momentum 0.9 and a learning rate of 0.01 for $K$ adaptation steps, with the values specified in Appendix~\ref{app:hyperparams}. The diffusion-based self-supervision is applied to timesteps $t \leq 100$, matching the training noise schedule. We set $T_s=0$ for all datasets to capture early propagation dynamics during training. Dataset-specific hyperparameter configurations are provided in Appendix~\ref{app:hyperparams}.

\begin{table}[pos=!tb]
\centering
\caption{Performance comparison on simulated propagation datasets. Best results are shown in bold, and second-best results are underlined. Improv. denotes the relative change of TGLR compared with the best baseline in each column.}
\label{tab:simulated_results}
\setlength{\tabcolsep}{4pt}
\csffittable{\begin{tabular}{lcccccccc}
\toprule
& \multicolumn{2}{c}{\textbf{Karate}} & \multicolumn{2}{c}{\textbf{Jazz}} & \multicolumn{2}{c}{\textbf{Cora-ML}} & \multicolumn{2}{c}{\textbf{Facebook}} \\
\cmidrule(lr){2-3} \cmidrule(lr){4-5} \cmidrule(lr){6-7} \cmidrule(lr){8-9}
\textbf{Method} & F1 & AUC & F1 & AUC & F1 & AUC & F1 & AUC \\
\midrule
\multicolumn{9}{l}{\textit{SI Model}} \\
LPSI~(2017)   & 0.348 & 0.876 & 0.290 & 0.872 & 0.448 & 0.867 & 0.398 & 0.834 \\
GCNSI~(2019)  & 0.499 & \underline{0.892} & 0.270 & 0.713 & 0.392 & 0.879 & 0.321 & 0.780 \\
IVGD~(2022)   & 0.505 & 0.883 & 0.464 & 0.836 & 0.482 & \underline{0.929} & 0.493 & 0.870 \\
SL-VAE~(2022)  & 0.211 & 0.536 & 0.144 & 0.509 & 0.011 & 0.501 & 0.026 & 0.503 \\
DDSL~(2024)   & 0.505 & 0.830 & 0.465 & 0.840 & 0.477 & 0.861 & 0.206 & 0.882 \\
SIDSL~(2025)  & 0.436 & 0.688 & 0.359 & 0.821 & 0.426 & 0.605 & 0.496 & \underline{0.913} \\
GDFSL~(2025)  & 0.421 & 0.762 & 0.235 & 0.692 & 0.239 & 0.718 & 0.194 & 0.684 \\
GTSI~(2025)   & \underline{0.527} & 0.889 & \underline{0.491} & \textbf{0.911} & \underline{0.525} & 0.925 & \underline{0.503} & 0.906 \\
PDSL~(2026)  & 0.358 & 0.584 & 0.219 & 0.528 & 0.166 & 0.501 & 0.169 & 0.505 \\
TGLR~(Ours)   & \textbf{0.543} & \textbf{0.927} & \textbf{0.498} & \underline{0.910} & \textbf{0.568} & \textbf{0.939} & \textbf{0.597} & \textbf{0.948} \\
\midrule
Improv. & +2.9\% & +3.9\% & +1.3\% & -0.1\% & +8.2\% & +1.0\% & +18.7\% & +3.8\% \\
\midrule
\multicolumn{9}{l}{\textit{SIR Model}} \\
LPSI~(2017)   & \underline{0.416} & 0.773 & 0.281 & 0.825 & 0.417 & 0.815 & 0.348 & 0.751 \\
GCNSI~(2019)  & 0.323 & 0.729 & 0.240 & 0.671 & 0.376 & 0.828 & 0.294 & 0.730 \\
IVGD~(2022)   & \textbf{0.487} & \textbf{0.824} & 0.437 & 0.798 & 0.432 & 0.839 & 0.426 & 0.772 \\
SL-VAE~(2022)  & 0.177 & 0.488 & 0.117 & 0.501 & 0.046 & 0.502 & 0.042 & 0.501 \\
DDSL~(2024)   & 0.399 & 0.707 & \underline{0.440} & 0.815 & 0.435 & 0.803 & \underline{0.431} & 0.788 \\
SIDSL~(2025)  & 0.358 & 0.625 & 0.320 & 0.491 & 0.355 & 0.613 & 0.366 & 0.765 \\
GDFSL~(2025)  & 0.332 & 0.706 & 0.215 & 0.681 & 0.178 & 0.674 & 0.205 & 0.670 \\
GTSI~(2025)   & 0.374 & \underline{0.788} & \textbf{0.463} & \underline{0.855} & \underline{0.458} & \underline{0.842} & 0.427 & \underline{0.796} \\
PDSL~(2026)  & 0.310 & 0.495 & 0.203 & 0.524 & 0.167 & 0.499 & 0.167 & 0.500 \\
TGLR~(Ours) & 0.405 & 0.764 & \underline{0.440} & \textbf{0.862} & \textbf{0.487} & \textbf{0.871} & \textbf{0.471} & \textbf{0.836} \\
\midrule
Improv. & -16.9\% & -7.3\% & -5.0\% & +0.8\% & +6.3\% & +3.4\% & +9.4\% & +5.0\% \\
\bottomrule
\end{tabular}}
\end{table}

\subsection{Performance Comparison (RQ1)}

Table~\ref{tab:simulated_results} and Table~\ref{tab:realworld_results} report the performance comparison across simulated and real-world datasets. TGLR is compared with nine baselines on simulated datasets and six baselines on real-world datasets. SL-VAE and IVGD are excluded from real-world evaluation due to their reliance on unavailable infection probability matrices, while PDSL is excluded because it assumes fixed node dimensions and consistent topology across samples, which does not hold for variable-size real-world cascades. Although these methods can be approximately adapted, such modifications introduce extra assumptions and substantially alter their original settings. Therefore, they are omitted for fair comparison.

\begin{table}[pos=!tb]
\centering
\caption{Performance comparison on real-world propagation datasets and the Twitter25-NQ variant without quotes.}
\label{tab:realworld_results}
\setlength{\tabcolsep}{4pt}
\csffittable{\begin{tabular}{lcccccccccc}
\toprule
& \multicolumn{2}{c}{\textbf{Twitter25}} & \multicolumn{2}{c}{\textbf{Twitter25-NQ}} & \multicolumn{2}{c}{\textbf{Twitter15}} & \multicolumn{2}{c}{\textbf{Twitter16}} & \multicolumn{2}{c}{\textbf{Weibo}} \\
\cmidrule(lr){2-3} \cmidrule(lr){4-5} \cmidrule(lr){6-7} \cmidrule(lr){8-9} \cmidrule(lr){10-11}
\textbf{Method} & F1 & AUC & F1 & AUC & F1 & AUC & F1 & AUC & F1 & AUC \\
\midrule
LPSI~(2017)   & 0.116 & 0.770 & 0.157 & 0.777 & 0.091 & 0.733 & 0.084 & 0.728 & 0.131 & 0.745 \\
GCNSI~(2019)  & \underline{0.317} & \underline{0.802} & 0.313 & \underline{0.815} & \underline{0.307} & \textbf{0.783} & \underline{0.307} & \underline{0.782} & \textbf{0.325} & \textbf{0.792} \\
DDSL~(2024)   & 0.275 & 0.692 & 0.265 & 0.672 & 0.279 & 0.719 & 0.276 & 0.726 & 0.279 & 0.713 \\
SIDSL~(2025)   & 0.251 & 0.511 & 0.220 & 0.632 & 0.228 & 0.705 & 0.212 & 0.703 & 0.216 & 0.644 \\
GDFSL~(2025)   & 0.142 & 0.693 & 0.177 & 0.591 & 0.117 & 0.714 & 0.100 & 0.714 & 0.146 & 0.636 \\
GTSI~(2025)   & 0.310 & 0.789 & \underline{0.360} & 0.796 & 0.292 & 0.774 & 0.279 & 0.754 & 0.306 & 0.759 \\
TGLR~(Ours)   & \textbf{0.342} & \textbf{0.807} & \textbf{0.380} & \textbf{0.818} & \textbf{0.314} & \underline{0.781} & \textbf{0.319} & \textbf{0.789} & \underline{0.311} & \underline{0.789} \\
\midrule
Improv. & +7.9\% & +0.6\% & +5.5\% & +0.4\% & +2.3\% & -0.3\% & +3.8\% & +0.9\% & -4.2\% & -0.3\% \\
\bottomrule
\end{tabular}}
\end{table}

\textbf{Under the SI model}, TGLR achieves the best F1 across all four datasets and the best AUC on three of them. The improvements are particularly substantial on large-scale networks: 18.7\% F1 gain on Facebook and 8.2\% on Cora-ML. GTSI, as the second-best performer in F1, demonstrates strong performance with Graph Transformer-based attention mechanisms but shows performance gaps on rich networks compared to TGLR. On smaller networks such as Karate and Jazz, TGLR maintains consistent superiority with more modest improvements of 2.9\% and 1.3\%, while GTSI remains competitive, reflecting that simpler network structures naturally reduce performance differentiation. TGLR also improves AUC on most networks, indicating better ranking quality.

\textbf{Under the SIR model}, TGLR achieves the best F1 on Cora-ML and Facebook, with substantial gains of 6.3\% and 9.4\%, respectively. GTSI remains a strong competitor in several SIR settings, indicating that global attention mechanisms are beneficial for capturing propagation patterns under recovery dynamics. On Karate-SIR, TGLR ranks third in F1, with IVGD achieving the best performance. This can be attributed to the combination of extreme data scarcity and the complexity of SIR dynamics. The limited network provides sparse training samples that make latent distributions difficult to learn, while recovery allows observed graph configurations to recur. Such recurrence is more likely on small graphs, making different propagation histories harder to distinguish from the final observation and consequently complicating source localization. In such constrained conditions, IVGD's inverse problem formulation shows advantages over learning-based methods. However, TGLR maintains competitive AUC performance in most SIR scenarios, indicating its ability to distinguish source nodes from non-source nodes under recovery dynamics.

\textbf{On real-world propagation datasets}, TGLR achieves the best F1 in four out of five settings (Twitter25, Twitter25-NQ, Twitter15, and Twitter16), with relative improvements ranging from 2.3\% to 7.9\%. On Weibo, TGLR achieves the second-best F1, trailing GCNSI by 4.2\%. For AUC metrics, TGLR achieves the best performance on Twitter25, Twitter25-NQ, and Twitter16, and the second-best on Twitter15 and Weibo. The comparison between Twitter25 and Twitter25-NQ examines method performance when quoting interactions are included or excluded. TGLR achieves the best F1 in both settings, supporting its effectiveness when this additional diffusion mechanism is considered.

Performance improvements are generally more pronounced on larger and more complex networks, such as Cora-ML and Facebook, compared to smaller networks like Karate and Jazz. This suggests that TGLR's approach scales effectively with network complexity. Notably, the generally lower performance observed across all methods on real-world datasets compared to simulated networks stems from the fundamental differences in propagation characteristics: real-world diffusion does not follow fixed parametric propagation models and exhibits substantial node heterogeneity in terms of influence, activity patterns, and connectivity. Overall, TGLR achieves an average relative F1 improvement of 3.1\% over the best-performing baselines across all thirteen experimental configurations, which comprise eight simulated and five real-world propagation settings, demonstrating its effectiveness across diverse datasets, network scales, and propagation models.

\subsection{Ablation Study (RQ2)}
\label{sec:ablation}

To understand the contribution of each component in TGLR, we conduct ablation studies on representative datasets under both SI and SIR models. TGLR is built upon three core mechanisms: temporally guided residual refinement, latent diffusion regularization, and test-time adaptation. For a fine-grained analysis, we further decompose temporally guided residual refinement into two factors: distributional latent modeling, which encodes node features as probabilistic latent distributions, and temporal guidance, which uses temporal propagation dynamics to guide residual refinement of the resulting distributions during training. Four variants are evaluated by removing these factors: distributional latent modeling (w/o dist), temporal guidance in residual refinement (w/o temp), latent diffusion regularization (w/o diff), and test-time adaptation (w/o TTA). Table~\ref{tab:ablation} presents the results.

\begin{table}[pos=tbp]
\centering
\caption{Ablation study on Karate and Cora-ML datasets.}
\label{tab:ablation}
\setlength{\tabcolsep}{4pt}
\csffittable{\begin{tabular}{lcccc|cccc}
\toprule
& \multicolumn{4}{c}{\textbf{Karate}} & \multicolumn{4}{c}{\textbf{Cora-ML}} \\
\cmidrule(lr){2-5} \cmidrule(lr){6-9}
& \multicolumn{2}{c}{SI} & \multicolumn{2}{c}{SIR} & \multicolumn{2}{c}{SI} & \multicolumn{2}{c}{SIR} \\
\cmidrule(lr){2-3} \cmidrule(lr){4-5} \cmidrule(lr){6-7} \cmidrule(lr){8-9}
\textbf{Variant} & F1 & AUC & F1 & AUC & F1 & AUC & F1 & AUC \\
\midrule
TGLR (full) & \textbf{0.543} & \textbf{0.927} & \textbf{0.405} & \textbf{0.764} & \textbf{0.568} & \textbf{0.939} & \textbf{0.487} & 0.871 \\
w/o dist & 0.522 & 0.914 & 0.371 & 0.752 & 0.532 & 0.935 & 0.472 & 0.867 \\
w/o temp & 0.494 & 0.918 & 0.399 & 0.759 & 0.488 & 0.912 & 0.446 & 0.868 \\
w/o TTA & 0.524 & 0.921 & 0.396 & 0.763 & 0.567 & \textbf{0.939} & 0.482 & 0.870 \\
w/o diff & 0.528 & 0.910 & 0.401 & 0.755 & 0.566 & 0.938 & 0.475 & \textbf{0.872} \\
\midrule
$\Delta$ dist & -3.7\% & -1.3\% & -8.2\% & -1.6\% & -6.3\% & -0.4\% & -3.1\% & -0.5\% \\
$\Delta$ temp & -9.0\% & -0.9\% & -1.4\% & -0.7\% & -14.1\% & -2.8\% & -8.4\% & -0.3\% \\
$\Delta$ TTA & -3.4\% & -0.6\% & -2.2\% & -0.2\% & -0.2\% & 0.0\% & -0.9\% & -0.1\% \\
$\Delta$ diff & -2.7\% & -1.8\% & -0.8\% & -1.3\% & -0.3\% & 0.0\% & -2.4\% & +0.1\% \\
\bottomrule
\end{tabular}}
\end{table}

Temporal guidance in residual refinement emerges as a particularly beneficial component among the ablated factors, with its removal causing consistent F1 degradation across all evaluated settings. On Cora-ML, F1 drops by 14.1\% under the SI model and 8.4\% under the SIR model, while the impact on Karate is more modest. This pattern suggests that temporal propagation dynamics serve as the core guidance for residual refinement of latent distributions, with their effectiveness scaling with network complexity. The latent distribution modeling consistently contributes to model performance across datasets. On Cora-ML, removing this component causes F1 degradation of 6.3\% on SI model and 3.1\% on SIR model, indicating its importance for learning expressive propagation representations in complex networks. Notably, its impact exceeds that of diffusion regularization on Cora-ML, suggesting that the probabilistic latent structure itself is an important mechanism for capturing propagation dynamics in data-rich scenarios. On Karate, the effect remains comparable, with F1 degradation of 3.7\% on SI model and 8.2\% on SIR model, indicating that distributional representations provide consistent benefits across different network scales.

Diffusion regularization exhibits scenario-specific contributions. On Karate-SI, removing diffusion regularization causes decreases of 2.7\% in F1 and 1.8\% in AUC, demonstrating its value for structuring the latent space under data scarcity. On Cora-ML, its removal causes F1 decreases of 0.3\% under the SI model and 2.4\% under the SIR model. These results suggest that diffusion regularization complements temporal guidance and distributional modeling by providing a learned prior for snapshot-based source localization.

Test-time adaptation shows modest impact across most settings, with F1 decreases of 2.2\% on Karate-SIR and less than 1\% on Cora-ML. The most noticeable impact appears on Karate-SI, with a 3.4\% F1 decrease and a 0.6\% AUC decrease when TTA is removed. These results support using the learned diffusion prior to refine snapshot-based representations at inference, yielding additional F1 gains beyond those obtained through training alone.

The ablation study highlights complementary contributions: temporal guidance in residual refinement and distributional modeling consistently improve F1, while diffusion regularization provides a learned latent prior and test-time adaptation uses this prior to further refine snapshot-based representations at inference. Together, these findings support TGLR's multi-component design for handling diverse source localization scenarios.

\subsection{Design Choice Validation (RQ3)}

We validate two key architectural decisions through controlled experiments: (1) the choice of latent-level residual refinement 
over feature-level concatenation for temporal integration, and (2) broadening the latent distribution to accommodate training-inference 
variability versus reducing the gap through explicit distribution alignment.

\subsubsection{\textbf{Temporal Integration Strategy}}
\label{sec:temproal_integration}

\begin{figure}[pos=tbp]
    \centering
    \includegraphics[width=0.62\columnwidth]{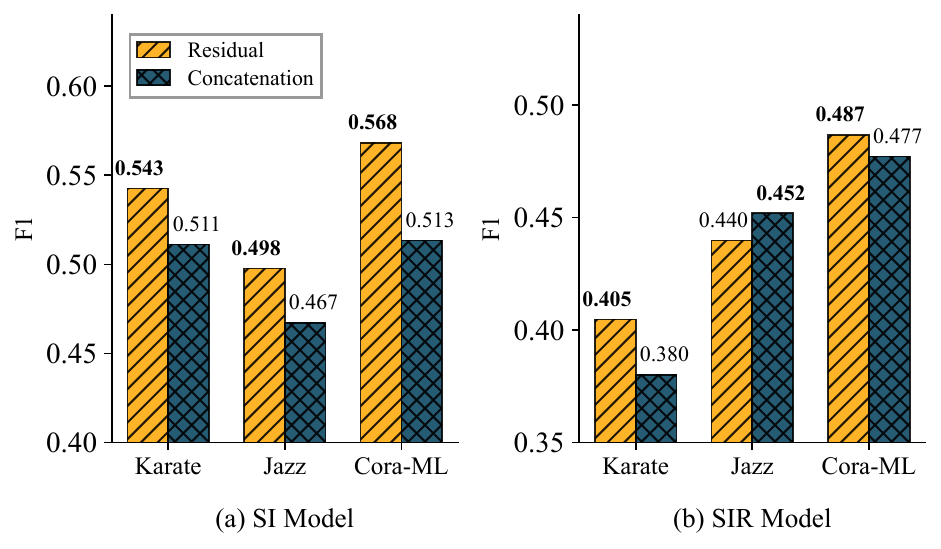}
    \caption{Performance comparison of temporal integration strategies under (a) SI and (b) SIR models.}
    \label{fig:design_integration}
\end{figure}

We compare our latent-level residual refinement approach against feature-level concatenation, where temporal context is directly concatenated with static features before encoding. Figure~\ref{fig:design_integration} presents the results across three representative datasets under both SI and SIR models.

Residual refinement achieves better performance than concatenation in most settings, with improvements ranging from 6.1\% to 10.7\% on SI models and up to 6.5\% on SIR models. The advantage is particularly pronounced on Jazz and Cora-ML, where residual refinement achieves up to 10.7\% 
improvement on Cora-ML under the SI model. Notably, under the SIR model, concatenation performs slightly better on Jazz. The consistent advantage on SI models validates our design rationale: by adding temporal context as a residual correction to the static representation, the model preserves base structural information while selectively incorporating dynamic patterns. In contrast, concatenation requires the encoder to jointly model static and temporal features, which makes it harder to preserve useful structural information while incorporating temporal dynamics.

\subsubsection{\textbf{Distribution Gap Handling Strategy}}
\label{sec:gap_strategy}

\begin{figure}[pos=tbp]
    \centering
    \includegraphics[width=0.62\columnwidth]{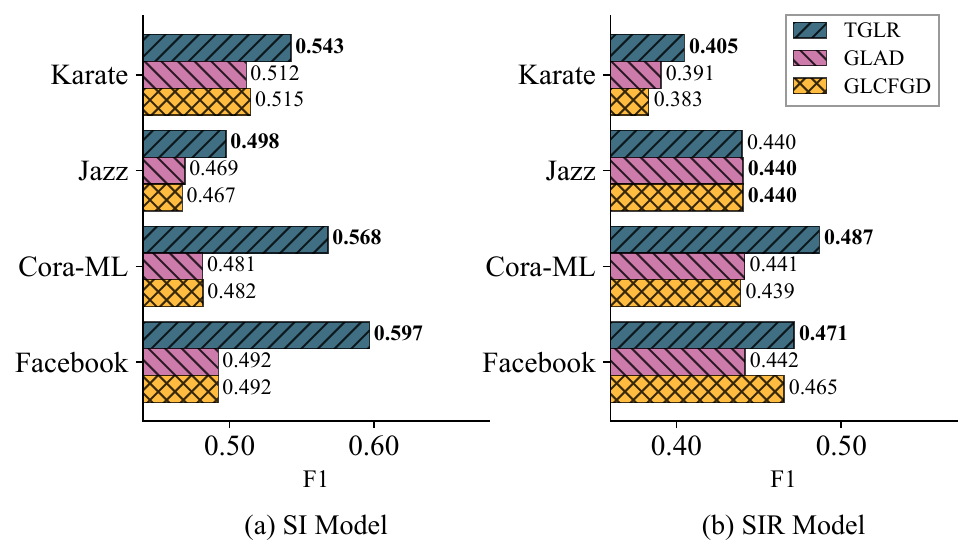}
    \caption{Performance comparison of distribution gap handling strategies under (a) SI and (b) SIR models.}
    \label{fig:design_distribution}
\end{figure}

To evaluate the impact of different strategies for handling the training-inference distributional gap, we compare three variants that share identical architectures up to and including temporally guided residual refinement. All three variants use the same encoder, VAE structure, and residual refinement mechanism to incorporate temporal propagation dynamics during training, while differing in how they address the distributional gap: (1) TGLR, which performs distribution expansion via diffusion regularization during training, followed by test-time adaptation during inference to adapt toward the learned distribution; (2) GLAD, which performs distribution alignment via knowledge distillation, where the student learns to mimic the teacher distribution refined by temporal context during training and directly uses the learned static distribution during inference; and (3) GLCFGD, which performs distribution alignment through classifier-free guidance, where the model is trained on both static-only and temporally refined representations, enabling prediction from static information alone during inference. Figure~\ref{fig:design_distribution} presents the comparison across four datasets.

TGLR achieves the best or competitive F1 performance across all datasets, with improvements ranging from 6.1\% to 21.3\% over GLAD and 5.4\% to 21.3\% over GLCFGD under SI models, and up to 10.2\% and 10.9\% respectively under SIR models. The performance gap is particularly pronounced on larger, more complex networks under the SI model. Interestingly, GLAD and GLCFGD achieve comparable performance in several settings despite using different alignment mechanisms, while TGLR generally achieves higher F1 scores, supporting the effectiveness of latent distribution refinement relative to the evaluated alignment strategies. Combined with the ablation findings from Section~\ref{sec:ablation}, these results suggest that although distributional modeling and temporal guidance in residual refinement provide important performance gains, retaining these shared components alone is insufficient: how the training–inference distributional gap is handled affects how effectively their representational capacity translates into source localization performance.

The similar performance of GLAD and GLCFGD indicates that both alignment strategies force the model to compress temporal information into a static-compatible distribution, inherently limiting distributional expressiveness. On complex networks with rich propagation dynamics, this compression discards nuanced temporal patterns, degrading performance despite the powerful base architecture. In contrast, TGLR allows the latent space to freely accommodate diverse observation states, fully leveraging the representational capacity of the base components. This validates that on complex networks, preserving distributional flexibility is more effective than enforcing distribution alignment when the base distributional modeling is sufficiently powerful.

\subsection{Hyperparameter Sensitivity Analysis (RQ4)}

\begin{figure}[pos=tbp]
    \centering
    \includegraphics[width=0.62\columnwidth]{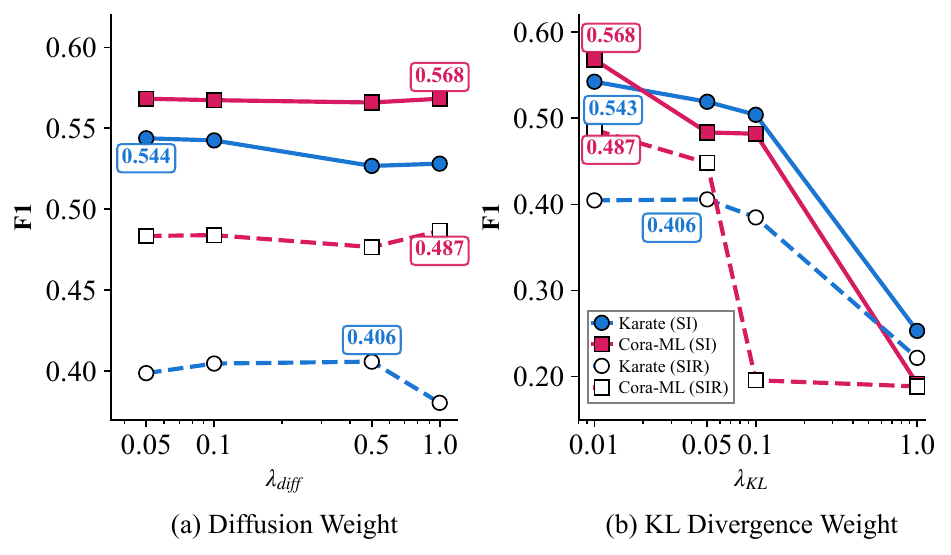}
    \caption{Hyperparameter sensitivity analysis: (a) Impact of diffusion regularization weight $\lambda_{\text{diff}}$; (b) Impact of KL divergence weight $\lambda_{\text{KL}}$.}
    \label{fig:hyperparameter_analysis}
\end{figure}

We analyze the sensitivity of TGLR to two critical hyperparameters: diffusion regularization weight $\lambda_{\text{diff}}$ and KL divergence 
weight $\lambda_{\text{KL}}$. Figure~\ref{fig:hyperparameter_analysis} presents performance variation across parameter settings on Karate 
and Cora-ML under SI and SIR models.

The diffusion weight $\lambda_{\text{diff}}$ shows dataset-dependent sensitivity. On Cora-ML, performance remains relatively stable across
$\lambda_{\text{diff}} \in [0.05, 1.0]$, with F1 varying by approximately 0.2 and 1.0 percentage points under SI and SIR, respectively, indicating robustness to the choice of diffusion weight within the evaluated range. In contrast, Karate exhibits greater sensitivity, with optimal performance
at $\lambda_{\text{diff}} = 0.05$ for SI and $\lambda_{\text{diff}} = 0.5$ for SIR. Performance degrades at $\lambda_{\text{diff}} = 1.0$, indicating that excessive
diffusion regularization can be detrimental on small networks with constrained latent space.

The KL divergence weight $\lambda_{\text{KL}}$ demonstrates critical sensitivity across all settings. Performance peaks at $\lambda_{\text{KL}} = 0.01$
in three of the four settings, while Karate-SIR achieves a slightly higher F1 at $\lambda_{\text{KL}} = 0.05$. Most strikingly, at $\lambda_{\text{KL}} = 1.0$, both Cora-ML and Karate suffer substantial performance degradation. At $\lambda_{\text{KL}} = 0.01$, the KL term provides gentle regularization that prevents latent space dispersion while preserving sufficient capacity for diverse propagation patterns. Higher weights constrain the posterior more strongly toward the prior, limiting expressiveness, with particularly pronounced performance degradation on Cora-ML. This supports our Section~\ref{sec:gap_strategy} finding that excessive regularization constraints can limit the benefits of the base architecture's representational capacity.

These findings summarize the sensitivity of TGLR to regularization weights, with Karate and Cora-ML serving as representative networks. They support using $\lambda_{\text{KL}} = 0.01$ across the evaluated settings. For $\lambda_{\text{diff}}$, values in [0.05, 0.1] are recommended for the evaluated small-scale networks, while the evaluated large-scale networks exhibit relatively low sensitivity within [0.1, 1.0].

\subsection{Distribution Refinement Visualization (RQ5)}

\begin{figure}[pos=tbp]
    \centering
    \includegraphics[width=0.95\textwidth]{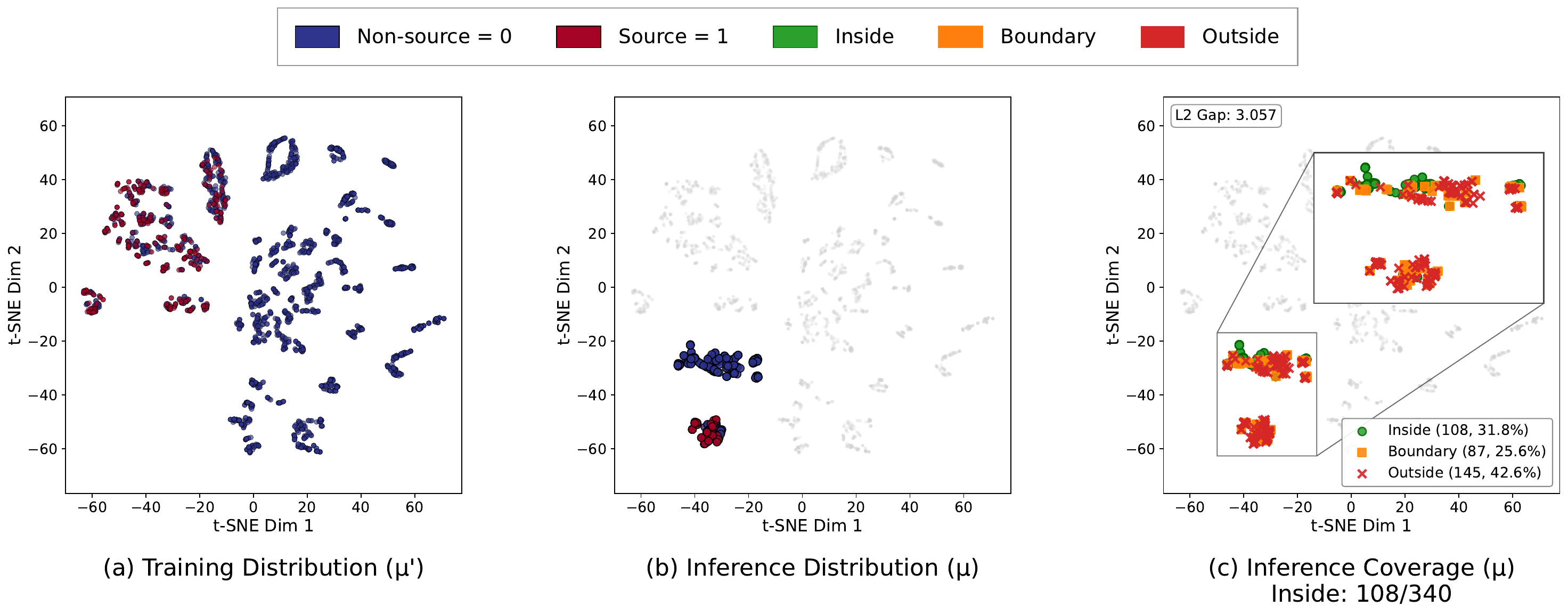}
    \caption{Visualization of the latent distribution shift induced by training-inference observation mismatch. 
    A baseline encoder without distributional bridging is used to isolate the effect of the mismatch: 
    training representations $\boldsymbol{\mu}'$ are derived from complete temporal sequences via residual refinement, 
    while inference representations $\boldsymbol{\mu}$ are encoded from static snapshots alone.
    Left to right: (a) Training distribution of $\boldsymbol{\mu}'$ learned from complete temporal sequences; 
    (b) Inference distribution of $\boldsymbol{\mu}$ encoded from static snapshots, overlaid on the training distribution; 
    (c) Coverage analysis with gap quantifying the distributional shift.
    }
    \label{fig:ablation_manifold}
\end{figure}

\begin{figure}[pos=tbp]
    \centering
    \includegraphics[width=0.90\textwidth]{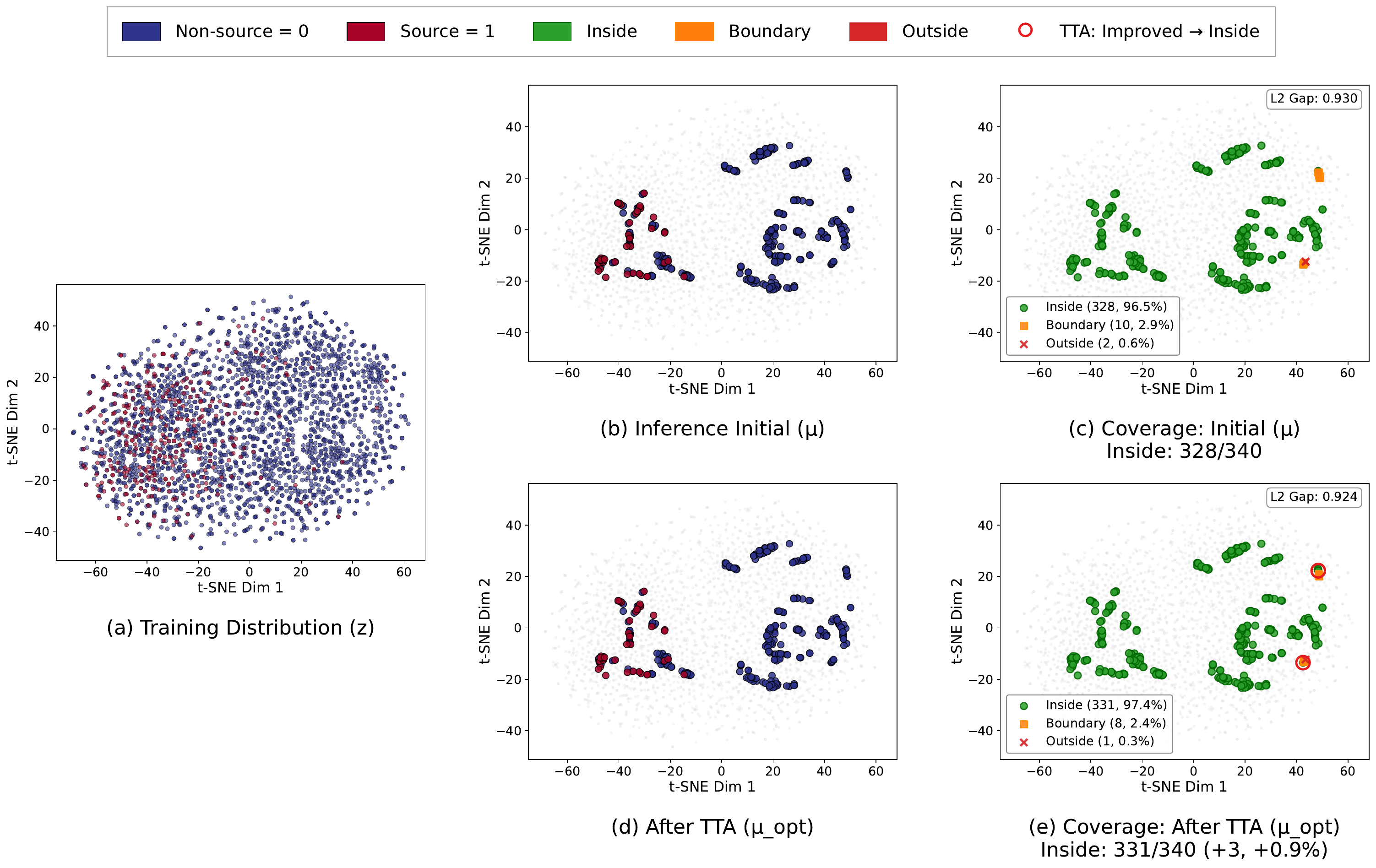}
    \caption{Distribution evolution visualization showing how test-time adaptation adapts static snapshot representations toward the 
    dynamics-enriched training distribution. Left to right: (a) Training distribution learned from complete temporal sequences; (b) Initial 
    inference representations ($\boldsymbol{\mu}$) encoded from static features; (c) Distributional coverage of initial inference representations against the training distribution; (d) Refined representations ($\boldsymbol{\mu}_{\text{opt}}$) after TTA optimization; (e) Distributional coverage after TTA optimization, with improvement in inside-distribution ratio annotated.}
    \label{fig:manifold_refinement}
\end{figure}

To directly observe the distributional effect of the training-inference observation mismatch, we employ a baseline encoder that applies residual refinement over temporal context during training but encodes only static snapshots at inference time, without any distributional bridging mechanism. Figure~\ref{fig:ablation_manifold} confirms the resulting distributional shift. The training representations $\boldsymbol{\mu}'$, derived from complete temporal sequences, form a structured latent distribution. When inference representations $\boldsymbol{\mu}$ are encoded from static snapshots alone, only 31.8\% fall within the high-density region of the training distribution, while 25.6\% reside in boundary areas and 
42.6\% fall entirely outside. The $L_2$ distance between training and inference centroids in latent space reaches 3.06, quantifying the magnitude of the distributional shift. These observations confirm that restricting inference to static snapshots while training on complete temporal sequences produces fundamentally misaligned representations, directly undermining generalization at inference time.

To provide interpretable insights into how test-time adaptation operates, we visualize the latent distribution evolution during inference. 
Figure~\ref{fig:manifold_refinement} illustrates the complete workflow: the training distribution learned from temporal sequences, initial inference representations obtained from static snapshots, their refinement through test-time adaptation, and quantification of distributional coverage before and after adaptation. Latent distribution modeling and diffusion regularization broaden the training distribution to accommodate diverse observation states at inference, while test-time adaptation actively adapts static-snapshot representations toward the dynamics-enriched training distribution. Compared with the baseline in Fig.~\ref{fig:ablation_manifold}(c), the initial inference representations in Fig.~\ref{fig:manifold_refinement}(c) exhibit a reduction in the out-of-distribution ratio from 42.6\% to 0.6\% and in the centroid distance from 3.06 to 0.93, reflecting the effects of latent distribution modeling and diffusion regularization before test-time adaptation. Latent distribution modeling combined with diffusion regularization first broadens the latent distribution, enabling 96.5\% of static-snapshot inference representations $\boldsymbol{\mu}$ to fall within the high-density region, with only 0.6\% of samples remaining outside the distribution. TTA optimization further refines the remaining boundary and outside cases, improving distributional coverage from 96.5\% to 97.4\%, reducing the boundary ratio from 2.9\% to 2.4\%, and reducing the outside ratio from 0.6\% to 0.3\%. On the validation set, this refinement improves AUC from 0.9046 to 0.9058 while maintaining the same F1 of 0.5313, indicating that TTA provides incremental improvements after the major distribution mismatch has been addressed.

Beyond distribution alignment, the learned latent representations exhibit desirable structural properties. Feature diversity analysis shows no dead dimensions, with all latent dimensions maintaining a standard deviation above $10^{-6}$ and an average standard deviation of 1.01. Furthermore, PCA projection captures only 10.01\% of the variance in the first two components, indicating that the learned representations preserve high-dimensional characteristics rather than collapsing into low-dimensional subspaces. Controlled experiments on the same Karate-SI validation set further demonstrate that restoring all distributional bridging components improves F1 from 0.5237 to 0.5313 and AUC from 0.8708 to 0.9058, confirming that the distribution refinement achieved by TGLR translates into measurable localization performance gains.

\subsection{Computational Efficiency Analysis (RQ6)}

\begin{figure}[pos=t]
    \centering
    \includegraphics[width=0.50\columnwidth]{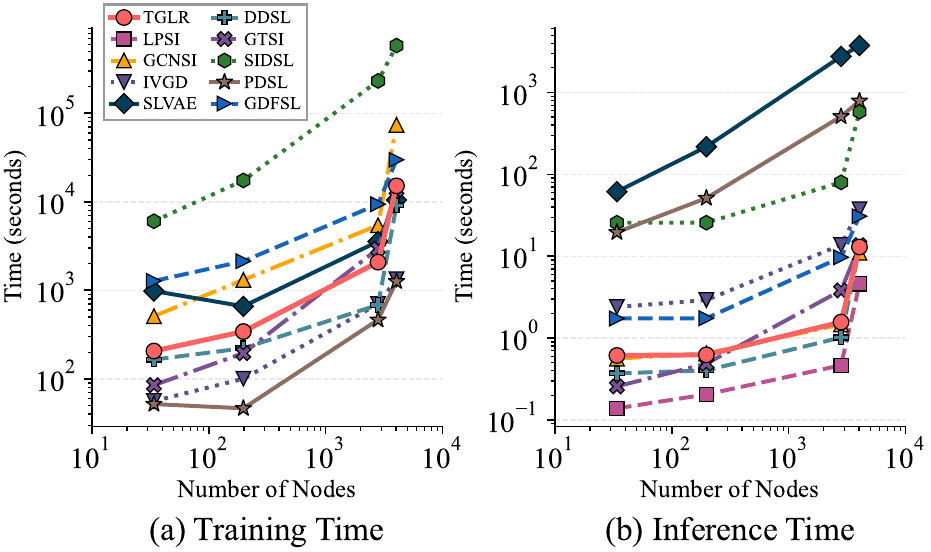}
    \caption{Computational efficiency comparison: (a) training time and (b) inference time.}
    \label{fig:efficiency_analysis}
\end{figure}

We evaluate the computational cost of TGLR in terms of both training and inference time, comparing against all baseline methods across four datasets of varying scales. Figure~\ref{fig:efficiency_analysis} presents the results on a logarithmic scale to accommodate the wide performance range.

TGLR demonstrates competitive efficiency across both training and inference phases. 
Training time is comparable to methods such as SL-VAE and DDSL, and substantially faster than SIDSL on large-scale networks. Note that LPSI, as a heuristic method, has no training phase. The training overhead mainly comes from graph aggregation, GRU-based temporal context encoding, and latent diffusion regularization. 
During training, temporal propagation sequences are used to guide residual refinement of latent distributions, which introduces additional computation compared with static-snapshot-only methods. Inference remains highly efficient because it operates solely on static snapshots without temporal propagation sequences, thereby avoiding the temporal context encoding used during training. Although both TGLR and SL-VAE employ inference-time optimization, they differ fundamentally: TGLR's TTA performs lightweight gradient-based optimization on low-dimensional latent representations, whereas SL-VAE optimizes node-level variables across the entire graph and iteratively runs the forward model to match observed infection states. 
Overall, TGLR achieves a favorable balance between accuracy and efficiency, with moderate training overhead and efficient static-snapshot inference.

\section{Conclusion}

This work studies source localization under a training--inference observation mismatch, where temporal sequences are available during training but not during inference. This mismatch induces a distributional shift, thereby limiting source localization performance. To mitigate this challenge, we propose TGLR, which integrates temporal dynamics into latent distributions and learns a diffusion prior to guide snapshot-based adaptation, thereby mitigating the training--inference distributional shift and enabling accurate localization without temporal sequence observations during inference. The constructed Twitter25 dataset incorporates quoting interactions, extending source localization evaluation to richer propagation structures. Experiments across thirteen simulated and real-world propagation settings show that TGLR achieves the best or competitive F1 performance across different settings, yielding an average relative F1 improvement of 3.1\% over baseline methods, with gains of up to 18.7\% in some settings. Further analysis shows that: (1) controlled comparisons support the advantages of latent distribution refinement over the evaluated distribution alignment strategies, particularly on larger networks; (2) TGLR achieves competitive computational efficiency, with moderate training overhead and inference that requires no temporal sequence processing.

Despite these results, several limitations remain. On extremely small networks with complex dynamics, data scarcity limits the effectiveness of learned representations. Distribution broadening and test-time adaptation also lack explicit control: broadening may extend beyond intended boundaries, while iterative adaptation may shift representations toward unintended regions. Future work could explore community membership at different scales to complement the existing structural features. Another direction is to introduce explicit constraints on distribution broadening and guidance for latent updates to improve refinement stability. In addition, temporally guided residual refinement introduces moderate training overhead on large-scale networks, motivating more scalable temporal modeling approaches that reduce computational cost while retaining the benefits of propagation dynamics.

\appendix

\section{Propagation Simulation Settings}

\begin{table}[pos=H]
\centering
\caption{Simulation parameters for propagation models on simulated datasets.}
\label{tab:simulation_params}
\csffittable{\begin{tabular}{lcc}
\toprule
\textbf{Parameter} & \textbf{SI Model} & \textbf{SIR Model} \\
\midrule
Base infection probability ($\beta$) & 0.05 & 0.05 \\
Degree bias coefficient & 1.00 & 1.00 \\
Recovery probability ($\gamma$) & -- & 0.02 \\
Initial infected ratio & 10\% & 10\% \\
Maximum infection scale & 30\% & 30\% \\
\bottomrule
\end{tabular}}
\end{table}

Table~\ref{tab:simulation_params} summarizes the key parameters used in our propagation simulations. To reflect realistic heterogeneity in diffusion dynamics, we apply degree-based scaling to infection probabilities. Specifically, the node-specific infection probability is computed as:
\begin{equation}
\beta_u = \beta \cdot \left[1 + b \cdot \left(d_u^{\text{norm}} - 0.5\right)\right],
\end{equation}
where $\beta$ is the base infection probability, $d_u^{\mathrm{norm}} \in [0,1]$ is the normalized degree of node $u$, and $b$ is the degree bias coefficient. The subtraction of $0.5$ centers the adjustment at the midpoint of the normalized degree range: nodes with $d_u^{\mathrm{norm}}=0.5$ retain the base infection probability $\beta$, while deviations from this midpoint introduce degree-dependent increases or decreases. This formulation introduces mild heterogeneity in infection probabilities across nodes.

\section{Hyperparameter Settings}
\label{app:hyperparams}

\begin{table}[pos=H]
\centering
\caption{Hyperparameter configurations for simulated and real-world propagation settings.}
\label{tab:hyperparams_sim}
\setlength{\tabcolsep}{3pt}
\csffittable{\begin{tabular}{@{}l*{13}{c}@{}}
\toprule
\multirow{2}{*}{\textbf{Parameter}} & \multicolumn{2}{c}{\textbf{Karate}} & \multicolumn{2}{c}{\textbf{Jazz}} & \multicolumn{2}{c}{\textbf{Cora-ML}} & \multicolumn{2}{c}{\textbf{Facebook}} & \multirow{2}{*}{\textbf{Twitter25}} & \multirow{2}{*}{\shortstack{\textbf{Twitter25-}\\\textbf{NQ}}} & \multirow{2}{*}{\textbf{Twitter15}} & \multirow{2}{*}{\textbf{Twitter16}} & \multirow{2}{*}{\textbf{Weibo}} \\
\cmidrule(lr){2-3} \cmidrule(lr){4-5} \cmidrule(lr){6-7} \cmidrule(lr){8-9}
& SI & SIR & SI & SIR & SI & SIR & SI & SIR & & & & & \\
\midrule
$\lambda_{\text{diff}}$ & 0.10 & 0.10 & 0.10 & 0.10 & 1.00 & 1.00 & 1.00 & 1.00 & 0.10 & 0.10 & 0.01 & 0.01 & 0.01 \\
$T_{\text{obs}}$ & 20 & 20 & 5 & 5 & 5 & 5 & 5 & 5 & 5 & 5 & 2 & 3 & 2 \\
Sampling & \xmark & \xmark & \xmark & \xmark & \cmark & \cmark & \cmark & \cmark & \xmark & \xmark & \xmark & \xmark & \xmark \\
$K$ & 50 & 50 & 100 & 50 & 30 & 50 & 30 & 30 & 100 & 80 & 100 & 100 & 80 \\
\bottomrule
\end{tabular}}
\end{table}
\suppressfloats

Sampling in Table~\ref{tab:hyperparams_sim} refers to prediction sampling during training. When disabled, predictions use the refined mean $\mu'$ rather than a sample from the latent distribution, while sampling is still performed for loss computation. The hyperparameter values are selected based on experimental performance for each setting.

\section{Latent Distribution Broadening Analysis}
\label{app:variance}

\begin{table}[pos=tbp]
\centering
\caption{Latent variance statistics comparing TGLR and its variant 
without diffusion regularization (w/o diff) across datasets and models. 
$\bar{\sigma}^2$ denotes mean variance across latent dimensions; 
Dim. $>1$ denotes the percentage of latent dimensions with 
variance exceeding 1.}
\label{tab:variance}
\setlength{\tabcolsep}{3pt}
\csffittable{\begin{tabular}{lcccccccc}
\toprule
& \multicolumn{2}{c}{\textbf{Karate-SI}} & \multicolumn{2}{c}{\textbf{Karate-SIR}} & \multicolumn{2}{c}{\textbf{Cora-ML-SI}} & \multicolumn{2}{c}{\textbf{Cora-ML-SIR}} \\
\cmidrule(lr){2-3} \cmidrule(lr){4-5} \cmidrule(lr){6-7} \cmidrule(lr){8-9}
\textbf{Variant} & $\bar{\sigma}^2$ & Dim. $>1$ & $\bar{\sigma}^2$ & Dim. $>1$ & $\bar{\sigma}^2$ & Dim. $>1$ & $\bar{\sigma}^2$ & Dim. $>1$ \\
\midrule
TGLR        & 1.0641 & 100.0\% & 1.0626 & 100.0\% & 1.7862 & 93.8\% & 1.8009 & 93.8\% \\
w/o diff    & 0.9999 & 46.9\% & 0.9999 & 43.8\% & 0.9740 & 6.2\% & 0.9749 & 3.1\% \\
\bottomrule
\end{tabular}}
\end{table}

Intuitively, training $f_\theta$ to reconstruct $\mathbf{z}_0$ from corrupted samples $\mathbf{z}_t = \sqrt{\bar{\alpha}_t}\mathbf{z}_0 + 
\sqrt{1-\bar{\alpha}_t}\boldsymbol{\epsilon}$ across varying noise levels encourages the encoder to populate a broader region of latent space: 
the denoising objective implicitly penalizes representations that are too concentrated, as $f_\theta$ must recover clean latents from an 
expanded noisy neighborhood around each $\mathbf{z}_0$, inducing larger posterior variance $\sigma^2_{\text{diff}} > \sigma^2$.
Table~\ref{tab:variance} empirically validates the distributional broadening effect of diffusion regularization in these independent controlled experiments. With diffusion regularization, TGLR maintains $\bar{\sigma}^2$ above 1.0 across all settings, with nearly all latent dimensions exceeding unit variance—100.0\% on Karate and 93.8\% on Cora-ML. Without diffusion regularization, $\bar{\sigma}^2$ falls below 1.0 across all settings and the proportion of dimensions exceeding unit variance reduces to as low as 3.1\%, consistent with the theoretical expectation of $\sigma_{\text{diff}}^2 > \sigma^2$.

\noindent\begin{minipage}{\linewidth}
\section{Structural Statistics of Real-World Datasets}
\label{app:realworld_stats}

Table~\ref{tab:realworld_structural_stats} reports the numerical values used to construct Figure~\ref{fig:radar}, before the axis normalization described in its caption.

\begin{table}[pos=H]
\centering
\caption{Structural statistics of real-world propagation datasets before normalization for Figure~\ref{fig:radar}.}
\label{tab:realworld_structural_stats}
\setlength{\tabcolsep}{4pt}
\csffittable{\begin{tabular}{lcccc}
\toprule
\textbf{Property} & \textbf{Twitter25} & \textbf{Twitter15} & \textbf{Twitter16} & \textbf{Weibo} \\
\midrule
Interaction Density & 1.126 & 1.005 & 1.006 & 1.003 \\
Avg Degree  & 2.152 & 2.013  & 2.012  & 1.987 \\
Density & 0.015       & 0.009     & 0.009     & 0.015     \\
Clustering & 0.038       & 0.008     & 0.008     & 0.003     \\
Gini Coeff. & 0.527       & 0.492     & 0.490     & 0.487     \\
\bottomrule
\end{tabular}}
\end{table}
\end{minipage}


\section*{CRediT authorship contribution statement}
\textbf{Dongpeng Hou:}
Writing -- review and editing, Software, Validation, Visualization.
\textbf{Junqing Zhao:}
Methodology, Software, Data curation,
Investigation, Validation, Formal analysis, Visualization,
Writing -- original draft.
\textbf{Chao Gao:}
Conceptualization, Writing -- review and editing, Resources, Project administration, Funding acquisition.
\textbf{Xianghua Li:}
Writing -- review and editing, Supervision.
\textbf{Zhen Han:}
Writing -- review and editing.
\textbf{Andrey Leonidov:}
Writing -- review and editing.
\textbf{A. M. Raigorodskii:}
Writing -- review and editing.

\section*{Funding}
This work was supported by the National Natural Science Foundation of China (Nos. 62471403, 62261136549), the Fundamental Research Funds for the Central Universities (Nos. G2024WD0151, D5000240309).


\section*{Declaration of competing interest}
The authors declare that they have no known competing financial interests or personal relationships that could have appeared to influence the work reported in this paper.

\section*{Data and code availability}
The code and datasets are publicly available at \url{https://github.com/DancinPuppet/TGLR}. To protect user privacy, user identifiers within each Twitter25 cascade and cascade identifiers have been replaced with dataset-specific indices. The public dataset retains the graph structures, numerical features, and labels required for the experiments, while excluding raw post text, media, usernames, and location information.

\section*{Declaration of generative AI and AI-assisted technologies in the manuscript preparation process}
During the preparation of this work, the authors used an AI-assisted language editing service to improve the language and readability of the manuscript. After using this service, the authors reviewed and edited the content as needed and take full responsibility for the content of the publication.

\bibliographystyle{elsarticle-num-csf}
\bibliography{IEEEabrv,myabrv,reference}

\end{document}